%% file: main.tex
\documentclass{article} 
\usepackage{iclr2026_conference,times}

\input{math_commands.tex}

\usepackage{hyperref}
\usepackage{url}
\usepackage{booktabs}       
\usepackage{amsfonts}       
\usepackage{nicefrac}       
\usepackage{microtype}      
\usepackage{xcolor}         
\usepackage{mathpazo}
\usepackage{graphicx}
\usepackage{natbib}
\usepackage{amsmath}
\usepackage{adjustbox}
\usepackage{multirow}
\usepackage{subcaption}
\usepackage{amssymb}
\usepackage{pifont}

\newcommand{\cmark}{\color{green}\ding{51}}%
\newcommand{\xmark}{\color{red}\ding{55}}%

\newcommand{\algname}{\textsc{VITA-QinYu }}
\newcommand{\algnamens}{\textsc{VITA-QinYu}}

\title{\algnamens: Expressive Spoken Language Model for Role-Playing and Singing}

\author{
QinYu Team \& VITA-Team\thanks{Full author list in contributions}
}

\newcommand\sourcecode[1]{\metadata[\raisebox{-0.1em}{\includegraphics[height=1em]{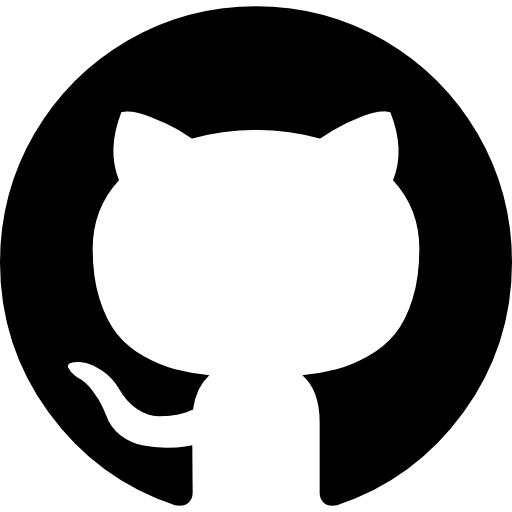}} Code]{\url{#1}}}
\newcommand\project[1]{\metadata[{\raisebox{-0.1em}{\includegraphics[height=1em]{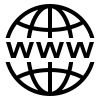}}} Project]{\url{#1}}}
\newcommand\demo[1]{\metadata[\raisebox{-0.2em}{\includegraphics[height=1em]{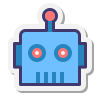}} Demo]{\url{#1}}}

\iclrfinalcopy 
\begin{document}

\maketitle

\begin{abstract}
    \input{sections/0.abstract}
\end{abstract}




\vspace{-1.5em}

\begin{center}
\small
\setlength{\tabcolsep}{0pt}
\begin{tabular}{ll}
\raisebox{-0.1em}{\includegraphics[height=1em]{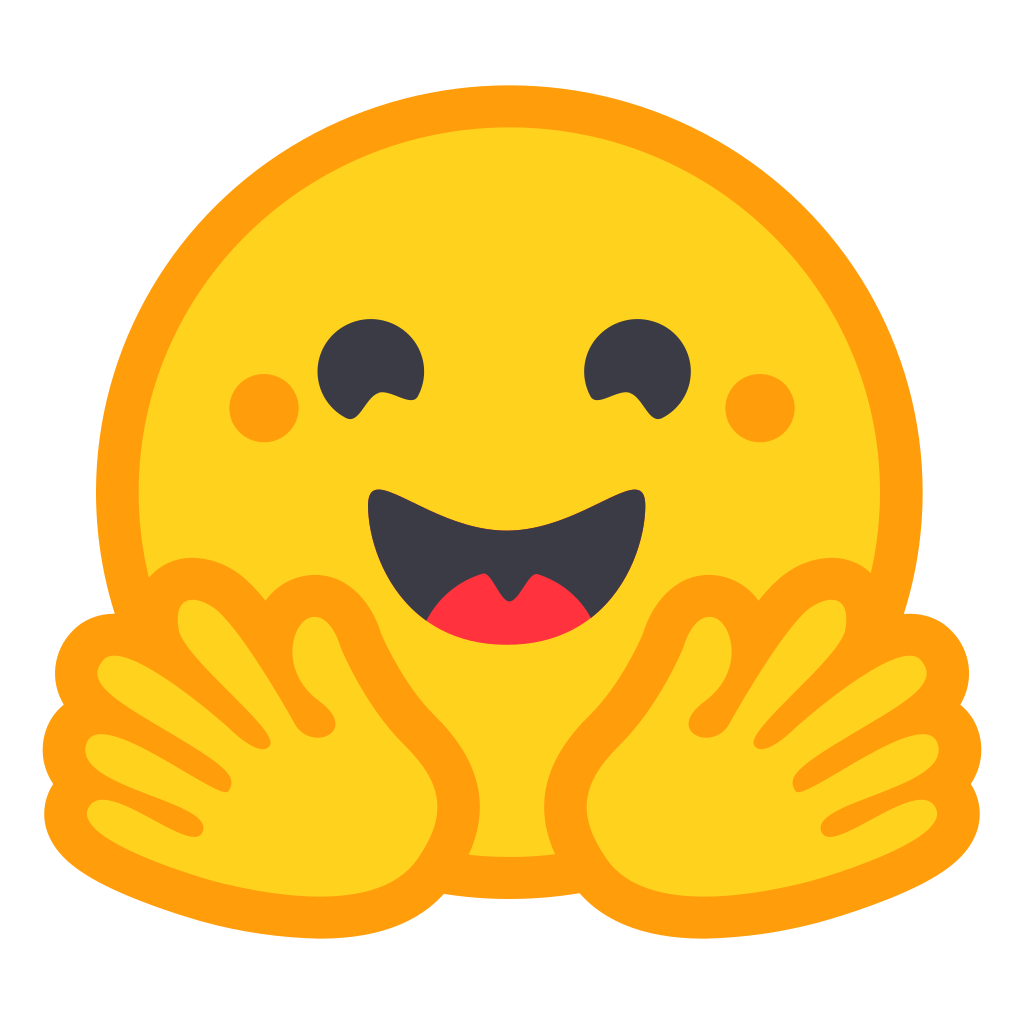}}~\textbf{Model:~} & \url{https://huggingface.co/collections/VITA-MLLM/vita-qinyu}\\
\raisebox{-0.1em}{\includegraphics[height=1em]{figures/github.png}}~\textbf{Code:~~~} & \url{https://github.com/VITA-MLLM/VITA-QinYu}\\
\raisebox{-0.1em}{\includegraphics[height=1em]{figures/icons8-website-100.png}}~\textbf{Demo:~~} & \url{https://tme-lyra-lab.github.io/VITA-QinYu}
\end{tabular}
\end{center}

\input{sections/1.intro}

\input{sections/2.related}

\input{sections/3.methods}

\input{sections/4.exp}

\input{sections/5.concl}

\subsubsection*{Author Contributions}

We would like to express our sincere gratitude to all contributors, including those not listed in the paper, for their invaluable support and efforts. The contributors are listed in no particular order. \\

{\bf Contributors}\\
Jiacheng Xu\textsuperscript{\rm 1}\quad 
Heting Gao\textsuperscript{\rm 2}\quad 
Liufei Xie\textsuperscript{\rm 1}\quad 
Zhenchuan Yang\textsuperscript{\rm 1}\quad 
Lijiang Li\textsuperscript{\rm 3}\quad 
Yiting Chen\textsuperscript{\rm 1}\quad 
Bin Zhang\textsuperscript{\rm 1}\quad 
Meng Chen\textsuperscript{\rm 1}\quad 
Chaoyu Fu\textsuperscript{\rm 3}\quad
Weifeng Zhao\textsuperscript{\rm 1}\quad
Wenjiang Zhou\textsuperscript{\rm 1}\quad


{\bf Affiliations}\\
\textsuperscript{\rm 1}TME Lyra Lab\quad 
\textsuperscript{\rm 2}Tencent YouTu Lab\quad 
\textsuperscript{\rm 3}Nanjing University

\bibliography{reference}
\bibliographystyle{iclr2026_conference}

\appendix
\section{Appendix}
\input{sections/append}

\end{document}

%% file: math_commands.tex
\usepackage{amsmath,amsfonts,bm}

\def\eqref#1{equation~\ref{#1}}

\def\1{\bm{1}}

\DeclareMathAlphabet{\mathsfit}{\encodingdefault}{\sfdefault}{m}{sl}
\SetMathAlphabet{\mathsfit}{bold}{\encodingdefault}{\sfdefault}{bx}{n}



%% file: sections/0.abstract.tex
Human speech conveys expressiveness beyond linguistic content, including personality, mood, or performance elements, such as a comforting tone or humming a song, which we formalize as role-playing and singing. We present \algnamens, the \emph{first} expressive end-to-end (E2E) spoken language model (SLM) that goes beyond natural conversation to support both role-playing and singing generation. 
\algname adopts a hybrid speech–text paradigm that extends interleaved text–audio modeling with multi-codebook audio tokens, a design enabling richer paralinguistic representation while preserving a clear separation between modalities to avoid interference.
We further develop a comprehensive data generation pipeline to synthesize a total of 15.8K hours of natural conversation, role-playing, and singing data for training. 
\algname demonstrates superior expressiveness, outperforming peer SLMs by $7$ percentage points on objective role-playing benchmarks, and surpassing peer models by $0.13$ points on a 5-point MOS scale for singing. 
Simultaneously, it achieves state-of-the-art conversational accuracy and fluency, exceeding prior SLMs by 1.38 and 4.98 percentage points on the C3 and URO benchmarks, respectively.
We open-source our code and models and provide an easy-to-use demo with full-stack support for streaming and full-duplex interaction.

%% file: sections/1.intro.tex
\section{Introduction}

End-to-end (E2E) spoken language models (SLMs) have achieved strong progress in fluent and informative conversational abilities, with performance in understanding, reasoning, and instruction following approaching text-only models~\citep{chen2025minmo,Zhang2025MiMoAudioAL}. However, human speech carries rich paralinguistic cues—such as prosody, intonation, rhythm, and style—that convey personality and emotion. For instance, users may seek comforting speech or soft humming in specific situations. We formalize these aspects as role-playing and singing, viewing them as key forms of speech expressiveness~\citep{huang2025step}, which remain underexplored in E2E SLMs.



Existing expressive speech systems are largely task-specific and do not support general conversational assistants. Role-playing systems~\citep{li2023chatharuhirevivinganimecharacter,wang2024rolellmbenchmarkingelicitingenhancing,zhang2025omnicharacterimmersiveroleplayingagents} typically adopt cascaded pipelines that combine LLM-based text generation with external speech synthesis. While modular, these approaches introduce significant engineering complexity due to their multi-component design. Traditional singing voice synthesis (SVS) methods rely on lyrics and musical scores~\citep{pan2026syntheticsingersreviewdeeplearningbased}, limiting their use in real-world interactions where users only provide song or singer names. This motivates a more general setting where singing must be generated from minimal natural-language inputs.

A comparison of recent LLMs and SLMs is shown in Table~\ref{tab:cap}. Motivated by these limitations, we propose \algname, the first E2E SLM supporting expressive speech generation alongside natural conversation. \algname adopts a hybrid speech–text paradigm, extending interleaved modeling~\citep{zeng2024scaling} with parallel multi-codebook audio token modeling~\citep{xie2024mini}, improving paralinguistic expressivity while reducing cross-modal interference~\citep{nguyen2025spirit}. As a native end-to-end system, it avoids the complexity of cascaded pipelines.

To support expressive generation, we construct large-scale datasets for role-playing and singing. Our 2.6K-hour role-playing dataset covers 20K+ roles, derived from audiobooks with structured character extraction and LLM-generated interactive scripts, followed by instruction-based expressive speech synthesis. We also build a 1.2K-hour singing dataset by collecting trending songs, using MIDI-guided zero-shot SVS for high-quality vocals, and converting song information into natural language instructions for conversational modeling.



\input{tables/capability}

We view role-playing and singing as early steps toward broader expressive speech generation. We hope this work provides a foundation for future research and will continue improving \algname for these capabilities.


Our contributions are summarized as follows:
\begin{itemize}
\item We propose \algname, the first E2E SLM with a hybrid text–speech paradigm supporting expressive role-playing and singing while maintaining strong conversational ability.
\item We construct 3.8K hours of role-playing and singing datasets to address gaps in expressive speech modeling.
\item Experiments show that \algname achieves strong expressiveness, outperforming prior SLMs on role-playing and singing benchmarks, while also matching or exceeding state-of-the-art conversational performance.
\end{itemize}

%% file: tables/capability.tex
\begin{table*}[t]
\caption{Comparison of existing LLM and SLMs on speech modaldity (Speech), natural conversation (Natural Conv.), role-playing (Role-Play), end-to-end architecture (Arch.) and speech-text modeling paradigm (Paradigm). ``N/A'' denotes ``not applicable''.}
\label{tab:cap}
\centering
\small
\begin{adjustbox}{max width=0.9\textwidth}
\begin{tabular}{lcccccccc}
\toprule
\textbf{Model} 
& \textbf{Speech}
& \textbf{Natural Conv.} 
& \textbf{Role-Play}
& \textbf{Singing}
& \textbf{Arch.}
& \textbf{Paradigm} 
\\
\midrule
Qwen2.5-7B-Inst
& \xmark & \cmark & \cmark & \xmark & N/A       & N/A \\
Qwen3-8B 
& \xmark & \cmark & \cmark & \xmark & N/A       & N/A \\
Qwen2.5-Omni
& \cmark & \cmark & \xmark & \xmark & Aligned & Parallel \\
Qwen3-Omni 
& \cmark & \cmark & \xmark & \xmark & Aligned & Parallel \\
Freeze-Omni 
& \cmark & \cmark & \xmark & \xmark & Aligned & N/A \\
Step-Audio 
& \cmark & \cmark & \xmark & \xmark & Aligned & Interleaved \\
GLM-4-Voice 
& \cmark & \cmark & \xmark & \xmark & Native  & Interleaved \\
Kimi-Audio 
& \cmark & \cmark & \xmark & \xmark & Native & Parallel \\\midrule
\algname            
& \cmark & \cmark & \cmark & \cmark & Native & Hybrid \\
\bottomrule
\end{tabular}
\end{adjustbox}
\vspace{-0.1in}
\end{table*}


%% file: sections/2.related.tex
\section{Related Works}

\textbf{Spoken Language Models (SLMs)}\quad
E2E SLMs can be categorized by architecture and modeling paradigm. Architecturally, they include native and aligned models~\citep{chen2025minmo}. Native SLMs~\citep{defossez2024moshi,xie2024mini,zeng2024glm,gao2025lucy,Long2025VITAAudioFI,Zhang2025MiMoAudioAL} use a single decoder-only Transformer for joint text–audio modeling, but struggle with modality gaps and limited pre-training. Aligned SLMs~\citep{fang2024llama,chen2025minmo,xu2025qwen2,xu2025qwen3omnitechnicalreport} adopt a “Thinker–Talker” two-stage design to preserve reasoning. Systems like Minmo~\citep{chen2025minmo} and Qwen-Omni~\citep{xu2025qwen2,xu2025qwen3omnitechnicalreport} decouple reasoning and speech generation, but rely on separate synthesis modules, often limiting paralinguistic expressivity.

From a modeling perspective, parallel models~\citep{defossez2024moshi,xie2024mini,chen2024slam,gao2025lucy,ding2025kimi,Zhang2025MiMoAudioAL} use multi-codebook audio tokens for richer acoustics but may weaken text–speech alignment~\citep{nguyen2025spirit}, while interleaved models~\citep{zeng2024glm,Long2025VITAAudioFI,li2025baichuan} alternate text and speech tokens for better linguistic consistency but rely on simpler audio representations and extra decoders for prosody. Extensions such as Baichuan-Audio~\citep{li2025baichuan} combine both ideas with more complex decoding pipelines. \algname simplifies this design by replacing the flow-matching decoder with lightweight MLP heads, encouraging more unified text–audio modeling.
\textbf{Audio Tokenizer}\quad
The architectural choice of an audio tokenizer determines the trade-off between reconstruction fidelity, paralinguistic expressiveness, and inference efficiency.
Residual Vector Quantization based decoders~\cite{defossez2024moshi,ye2025codec,wang2025spark,siuzdak2024snac,gong2025xy} represent audio
through multiple codebooks. These codebooks naturally capture rich paralinguistic information, such as speaker identity and prosody. Since the representation is highly descriptive, it places less computational demand on the decoder; a simple CNN-based decoder is often sufficient to reconstruct high-quality audio with low latency.
In contrast, models like CosyVoice2~\citep{du2024cosyvoice} and GLM-4-Voice~\citep{zeng2024glm} rely on single-codebook semantic tokens. While these tokens are highly compressed for semantic efficiency, they often lead to the loss of paralinguistic details. In preliminary experiments, we find that these tokenizers fail to reconstruct the melody of the original singing voice. 

\textbf{Role-Playing Models}\quad
Recent advances in LLMs enable strong role-playing capabilities~\citep{chen2024persona}, enabling immersive character simulation. However, most speech role-playing systems remain cascaded. For example, ChatHaruhi~\citep{li2023chatharuhirevivinganimecharacter} generates role-consistent text via an LLM and relies on external TTS for speech. OmniCharacter~\citep{zhang2025omnicharacterimmersiveroleplayingagents} encodes user queries with Whisper~\citep{radford2023robust}, aligns them with a Qwen2.5-7B-Instruct~\citep{yang2024qwen2} backbone to produce text, then uses a separate speech LLM and synthesis module to generate role-aware speech.


\textbf{Singing Voice Synthesis Models}\quad
Traditional singing voice synthesis (SVS) generates high-fidelity vocals from lyrics and scores~\citep{pan2025synthetic}, with recent advances improving quality and modeling. VISinger~\citep{zhang2022visinger}, based on VITS~\citep{kim2021conditional}, enables end-to-end SVS; Toksing~\citep{wu2024toksing} uses a non-autoregressive LM over quantized representations; and HiddenSinger~\citep{hwang2025hiddensinger} applies latent diffusion on neural codecs. However, most SVS systems depend on structured inputs (e.g., MIDI), limiting use in interactive settings where users provide only natural language.

%% file: sections/3.methods.tex
\section{Methods}

\begin{figure}[t!]
    \begin{center}
    \includegraphics[width=0.95\columnwidth,trim={0.6cm 0.6cm 0.6cm 0.3cm},clip]{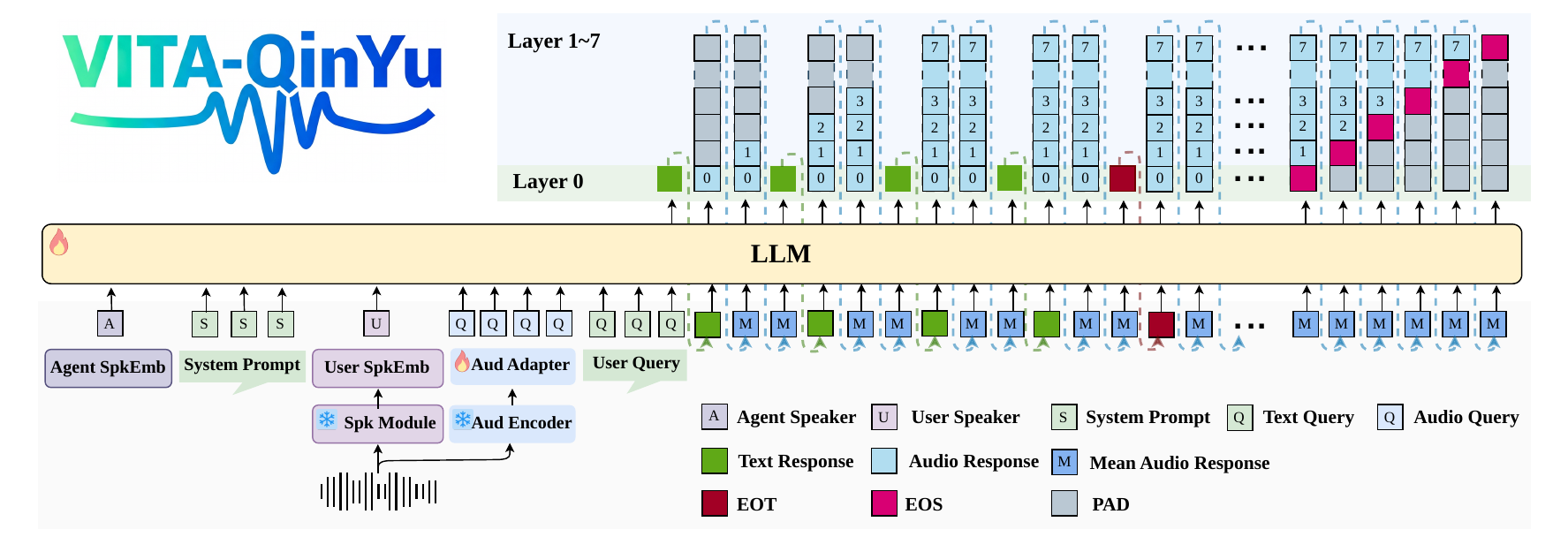}
    \caption{Architecture overview of \algnamens. 
    For text input, the LLM directly consumes embeddings; for speech input, a speaker module extracts speaker embeddings and an audio encoder extracts continuous features. An additional agent speaker embedding controls response timbre. Conditioned on these signals, the LLM generates interleaved text and multi-codebook audio tokens. Audio tokens are temporally shifted for quality, averaged back into the model for the next step, and decoded into waveforms. During training, the speaker and audio encoders are frozen, while only the adapters and LLM are updated.
    }
    \label{fig:arch}
    \end{center}
\end{figure}

\begin{figure}[ht]
    \centering
    \begin{subfigure}{0.585\textwidth}
        \includegraphics[width=\textwidth,trim={0.6cm 0.cm 0.7cm 0.3cm},clip]{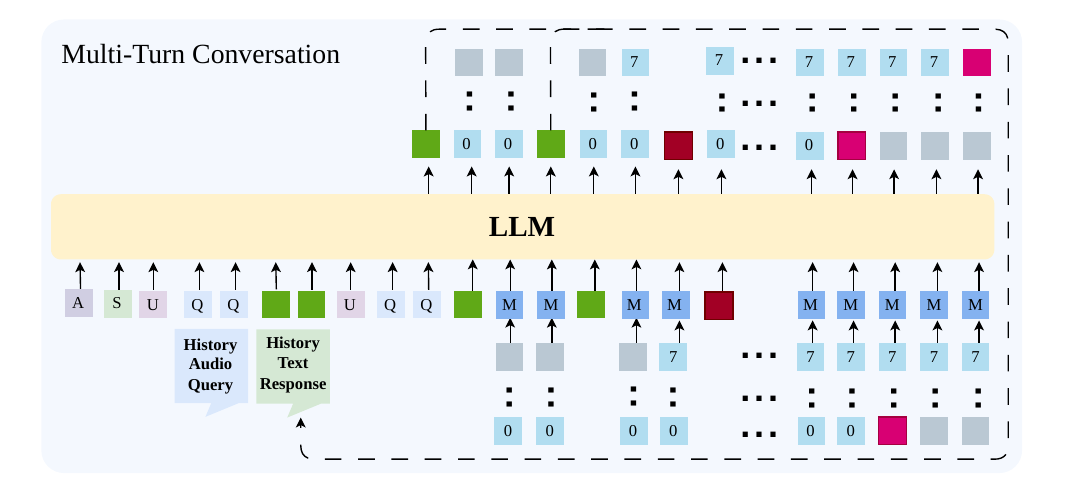}
        \caption{Multi-turn conversation.}
        \label{fig:mt}
    \end{subfigure}
    \hfill
    \begin{subfigure}{0.197\textwidth}
        \includegraphics[width=\textwidth,trim={0.6cm 0.7cm 0.65cm 1.cm},clip]{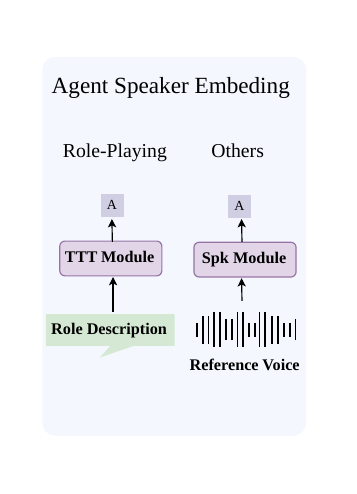}
        \caption{Agent speaker.}
        \label{fig:spk}
    \end{subfigure}
    \hfill
    \begin{subfigure}{0.197\textwidth}
        \includegraphics[width=\textwidth,trim={0.7cm 0.7cm 0.75cm 1.cm},clip]{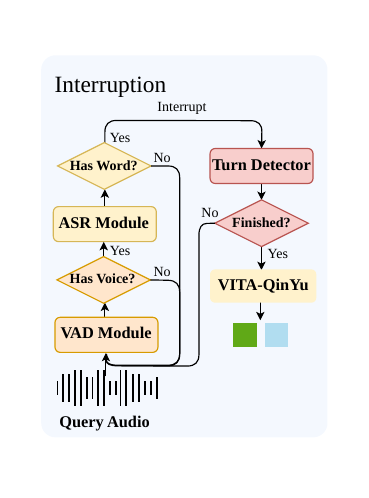}
        \caption{Interruption.}
        \label{fig:int}
    \end{subfigure}
    \caption{Logic of multi-turn conversation, agent speaker generation and interruption. 
    }
    \label{fig:submodule}
\end{figure}

The overview of \algnamens's architecture is shown in Figure~\ref{fig:arch},
which consists of an audio encoder, an audio adapter, a speaker embedding module, a language model backbone, and eight language-modeling heads. Additionally, a text-to-timbre (TTT) module is integrated into the system for role-playing tasks.
Detailed descriptions of each component are provided in the following sections.

\textbf{Backbone Model}\quad 
The backbone of \algname is a decoder-only Transformer based language model (LM). We experiment with Qwen3-8B~\citep{hu2026qwen3} and Youtu-LLM-4B~\citep{lu2025youtu}, resulting in two variants: \algnamens-8B and \algnamens-4B.
The backbone model processes users' queries, whether in speech or text, and generates both text and audio responses in a hybrid paradigm, which we formulate as follows.

Denote the user's input as $X\in\mathcal{X}$, where $\mathcal{X}$ is the joint space for text and speech embeddings. Denote the model's text response and speech response as $Y\in\mathcal{V}$ and $Z\in\mathcal{U}$ respectively, where $\mathcal{V}$ is the text vocabulary set and $\mathcal{U}$ is the speech codec vocabulary set. In the multi-codebook setting with $J$ codebooks, we have $\mathcal{U}=\cup_{j=0}^{J-1}\mathcal{U}^j$ and the speech tokens $Z$ can be multi-codebook tokens stacked in parallel: $Z=[Z^j]_{j=0}^{J-1}$, where the speech token $Z^j\in\mathcal{U}^j$ belongs the $j$-th codebook vocabulary set $\mathcal{U}^j$.
We interleave the text and speech response tokens with a predefined ratio of $n:m$ into a new sequence $S$ as follows:
\begin{equation}
    S=[Y_{0:n-1},Z_{0:m-1},Y_{n:2n-1},Z_{m:2m-1},\dots],
\end{equation}
where the text tokens and speech tokens are alternated in blocks of size $n$ and $m$, respectively. Once the text tokens are consumed, the remaining speech tokens are appended to the end of the sequence.
Denote the dataset as $\mathcal{D}=\{(X_i,S_i)\}_{i=1}^D$, where $D$ is the number of samples in the dataset. The negative log-likelihood $\mathcal{L}$ over the dataset $\mathcal D$ can be modeled as:
\begin{align}
    \mathcal{L}=&\sum_{i=1}^D\sum_{t=1}^{T_i}\log P(S_{i,t}|X_i,S_{i,<t})
\end{align}
where $T_i$ is the length of the interleaved sequence $S_i$. When $S_t\in\mathcal{V}$ is a text token, we compute the conditional log-probability the same as in the conventional LLM approach. When $S_t\in\mathcal{U}$ is the stacked speech tokens, the log-probability is modeled using the average log-probability of speech tokens across $J$ codebooks. Formally, the log-probability $\log P(S_t|X,S_{<t})$ is computed as:
\begin{equation}
    \log P(S_t|X_{<t},S_{<t})=\begin{cases} \log P(Y|X,S_{<t}), & \text{if $S_t$ is text: $S_t=Y$,} \\
    \frac{1}{J}\sum_{j=0}^{J-1}\log P(Z^j|X,S_{<t}) & \text{if $S_t$ is speech: $S_t=[Z^j]_{j=0}^{J-1},$}
    \end{cases}
\end{equation}
where the subscript $i$ is omitted for the sake of clarity.

\textbf{Multi-Turn Conversation}\quad  We prepend the conversation history to the LLM's input to support multi-turn interactions. The user's query, whether in text or speech, is included as-is. Since speech responses are often lengthy and largely redundant with the corresponding text responses, we discard the speech and retain only the text response in the history context, as illustrated in Figure~\ref{fig:mt}.


\textbf{Audio Encoder and Adapter}\quad 
We use SenseVoiceSmall~\citep{an2024funaudiollm} to encode input speech into $16.7$ Hz continuous features with a hidden size of $560$ and an MLP-based adapter to align the SenseVoiceSmall's output with the backbone language model. We keep the audio encoder frozen and train only the adapter. 

\textbf{Speech Decoder}\quad 
Following parallel models~\citep{xie2024mini, gao2025lucy}, we adopt the multi-codebook XY-Tokenizer~\citep{gong2025xy}, which encodes speech into eight 12.5 Hz codebooks (100 Hz total). It outperforms single-codebook tokenizers (e.g., CosyVoice~\citep{du2024cosyvoice2}, GLM-4-Voice~\citep{zeng2024glm}) in reconstructing both speech and singing. We average tokens per frame as LLM input, extend the LM head to predict text or the first audio layer, and add seven heads for the remaining layers. We find that a one-token delay per layer (Figure.~\ref{fig:arch}) improves speech quality; without it, speech is intelligible but overly fast.

\textbf{Speaker Encoder and Timbre Generation}\quad
LLM backbones can learn speaker embeddings for better timbre generalization~\citep{zhou2025indextts2, du2025cosyvoice, hu2026qwen3}. To enable diverse timbre control, we inject CAM++~\citep{wang2023cam++} speaker embeddings into the LLM during training (Figure.~\ref{fig:spk}). We compute averaged agent embeddings from 100 samples per voice and extract user embeddings from queries. At inference, we adopt the Text-to-Timbre module from DeepDubbing~\citep{dai2025deep}, which uses CFM~\citep{lipman2023flow} to generate agent embeddings from character descriptions.

\textbf{Full-Duplex Interaction}\quad
We use SileroVAD~\citep{SileroVAD} for voice activity detection and Whisper~\citep{radford2023robust} for ASR. Non-empty transcripts are processed by the turn detector TEN~\citep{TEN_Turn_Detection}; when it signals ``Finished,'' audio is sent to \algname for response generation, otherwise the system continues listening (Figure.~\ref{fig:int}).

\section{Data Collection}\label{sec:data}

\subsection{Pretraining Data}\label{sec:pretraining_data}
Following~\cite{long2025vita}, we collect open-source ASR, TTS, SQA, and text data for pretraining, and further augment it with in-house datasets. Details are provided in the Appendix.

\subsection{Conversational Data}
\begin{figure}[ht]
    \centering
    \begin{subfigure}{0.65\textwidth}
        \includegraphics[width=\textwidth,trim={5cm 3.2cm 7cm 2cm},clip]{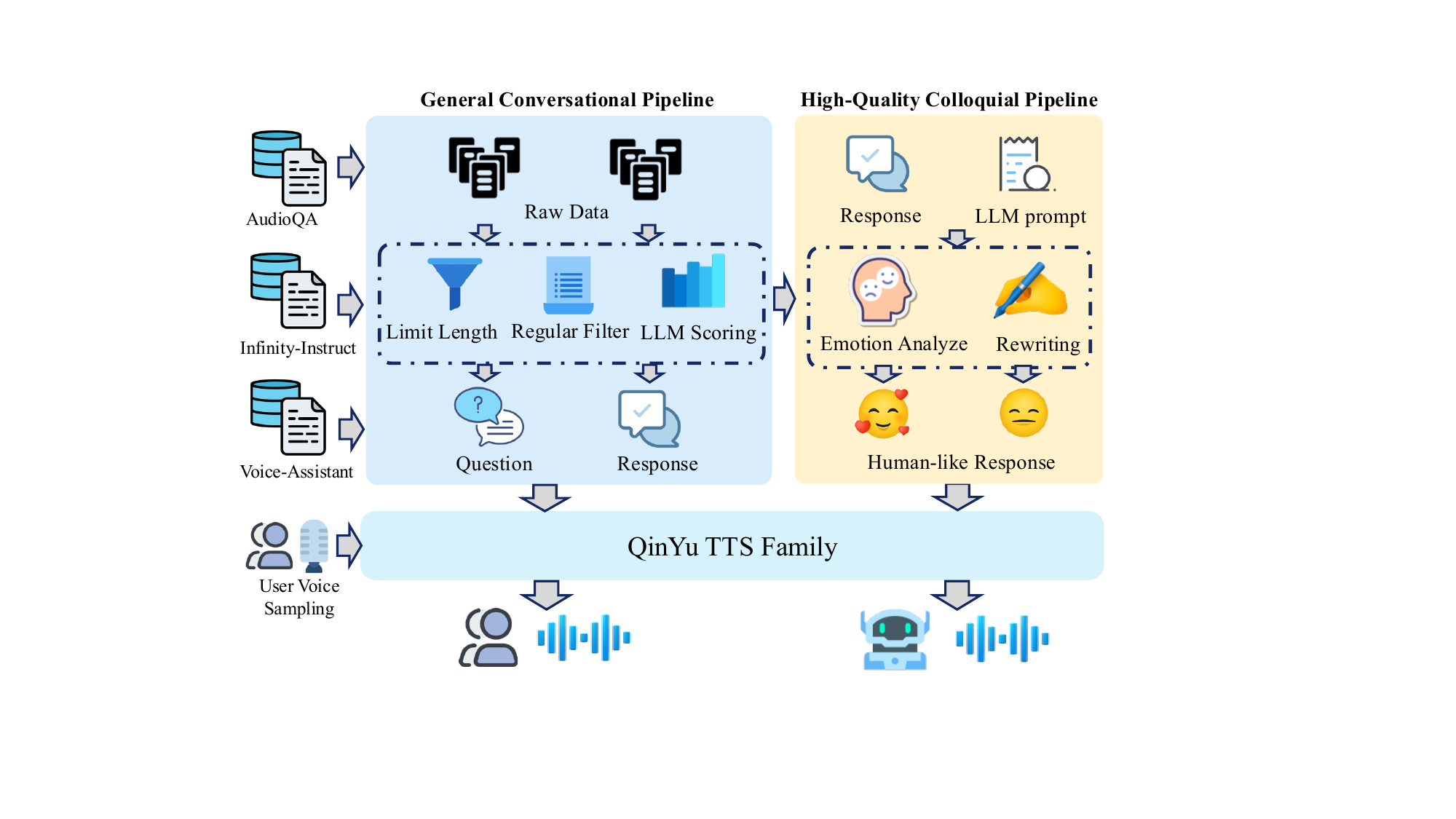}
        \caption{Conversational Data Pipeline.}
        \label{fig:pipe_conversation}
    \end{subfigure}
    \hfill
    \begin{subfigure}{0.55\textwidth}
        \includegraphics[width=\textwidth,trim={7cm 3.5cm 7cm 1.2cm},clip]{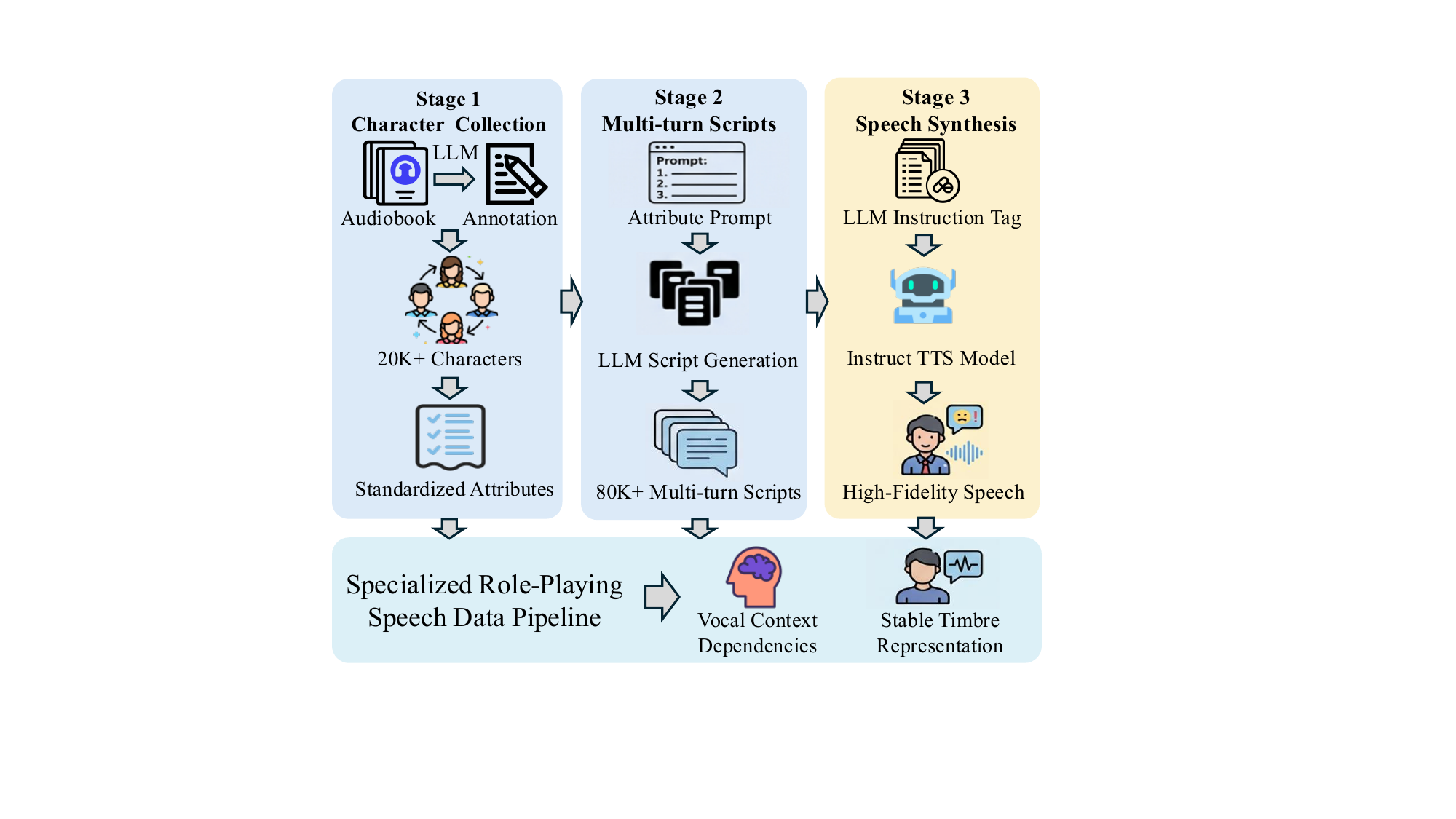}
        \caption{Role-Playing Data Pipeline. }
        \label{fig:pipe_roleplay}
    \end{subfigure}
    \hfill
    \begin{subfigure}{0.39\textwidth}
        \includegraphics[width=\textwidth,trim={10.6cm 4cm 10cm 2cm},clip]{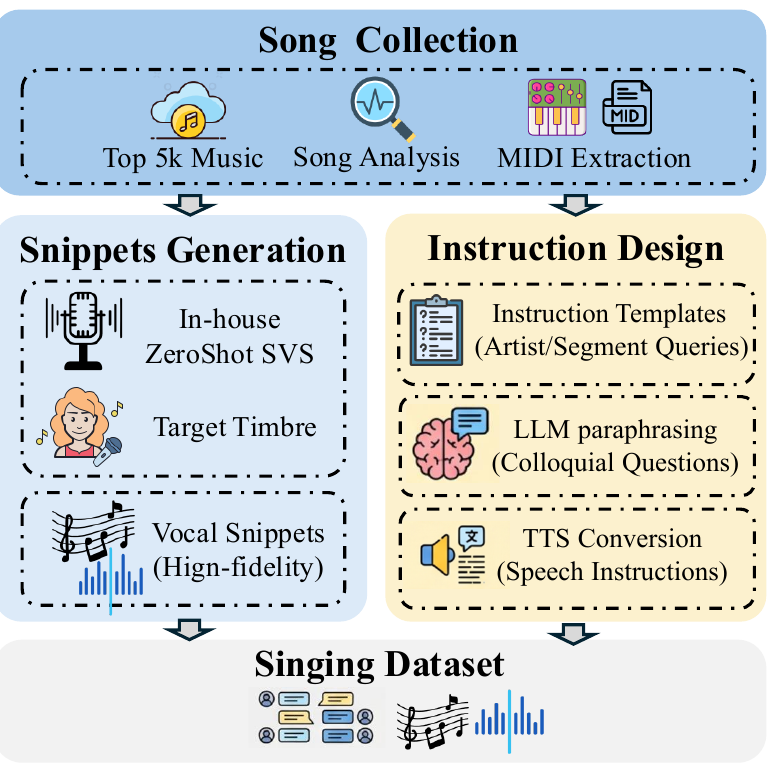}
        \caption{Singing Data Pipeline. }
        \label{fig:pipe_singing}
    \end{subfigure}
    \caption{Data pipelines for natural conversation, role-playing and singing.
    }
    \label{fig:data_pipeline}
\end{figure}

To endow our model with both fundamental conversational skills and more natural, colloquial expression patterns, we constructed a General Conversational Dataset and a High-Quality Colloquial Speech Dataset. The conversation data pipeline is illustrated in Figure~\ref{fig:pipe_conversation}.

\textbf{General Conversational Dataset}\quad We source raw text data from AudioQA-1.0M ~\citep{gao2025lucy}, Voice Assistant 400K ~\citep{xie2024mini}, and Infinity-Instruct ~\citep{li2025infinity}. To ensure data quality, we restricted the maximum token length to 800 to exclude excessively long responses. Simultaneously, we filter out content deviating from conversational scenarios, such as complex mathematical formulas and code snippets. For each response turn, we utilize DeepSeek to evaluate the content based on topic relevance, tone, and fluency. Scores were assigned on a scale of 0 to 1, and low-scoring dialogues were discarded. 
Ultimately, we curated 800K Chinese and English samples from AudioQA-1.0M, 300K Chinese samples from Infinity-Instruct, and 450K English samples from Voice Assistant 400K.

\textbf{High-Quality Colloquial Dataset}\quad Although current E2E SLMs support basic interaction, their responses are often unnatural due to limited prosodic variation. To address this, we select high-scoring samples from our general conversational dataset and refine the responses. Inspired by OpenS2S ~\citep{wang2025opens2s}, we place a strong emphasis on the emotional nuances inherent in human conversation and the flexibility of emotional expression. We leverage DeepSeek~\cite{liu2024deepseek} to analyze potential underlying emotions within the dialogue text and generate corresponding colloquial expressions based on these emotions, thereby further enhancing the human-like quality of the interactions. Consequently, we collect and rewrite 400K textual dialogue samples featuring diverse emotional expressions.

User speech input is naturally diverse, influenced by factors such as age, gender, and accents. To improve model robustness, we simulate realistic user queries by extracting over 90K unique speaker prompts from ASR datasets (e.g., Aishell and Common Voice). These queries are synthesized using an in-house zero-shot TTS model. For the response side, we utilize a speaker fine-tuned TTS model to ensure high-quality audio output.

\subsection{Role-playing Data}


The role-playing pipeline synthesizes high-fidelity multi-turn dialogs, enabling models to capture vocal context and maintain stable timbre. As shown in Figure~\ref{fig:pipe_roleplay}, it has three steps.
\textbf{(1)~Character Profiles Collection and Formalization}:
Despite often deviating from the interactive dialog structures, audiobooks contain diverse characters with distinct personalities, making them an ideal source of character profiles. 
Inspired by Deep Dubbing~\citep{dai2025deep}, we use LLMs to annotate large-scale audiobook corpora, producing a diverse library of over 20K unique character profiles, each associated with a distinct vocal timbre and defined by four key attributes: role demographics, social identity, behavioral temperament, and acoustic vocal style.
\textbf{(2) Character Multi-turn Script Construction}:
We design an attribute-constrained prompting strategy to construct high-quality role-playing scripts. Our generation prioritizes the alignment of persona, scenario, and linguistic style. 
For example, a character defined as a ``gentle snack shop owner'' may be instantiated in a ``casual purchase conversation.'' Specialized prompts based on personality descriptions are employed to ensure that generated responses remain consistent with the specified gentle temperament.
To support multi-turn interactions, each session includes 8–15 dialogue turns. This process yields a corpus of 80K role-playing scripts spanning diverse scenarios.
\textbf{(3) Instruction-Based Expressive Speech Synthesis}:
A key challenge in speech synthesis is the expressivity gap--mapping implicit text sentiment to explicit prosody. To address this, we use LLMs to add instruction tags to role-playing scripts and apply in-house instruct TTS models inspired by \cite{yang2024instructtts,du2025cosyvoice,zhou2025indextts2} to generate role-playing speech with fine-grained prosodic and emotional control.

\subsection{Singing Data}



Most SVS systems~\citep{pan2026syntheticsingersreviewdeeplearningbased} are score- or melody-controlled, limiting seamless human–agent singing interaction. To enable unconstrained interaction, we propose a scalable three-step pipeline for constructing the instruction-based singing dataset, as shown in Figure~\ref{fig:pipe_singing}.
\textbf{(1) Song Collection}:
To ensure our model captures a comprehensive distribution of contemporary vocal interactions, we curate a diverse corpus of 5K popular musical compositions. We perform structural decomposition on these tracks to identify fine-grained segments—such as verses and choruses. This temporal partitioning facilitates precise and localized control over the generative process. Subsequently, we extract symbolic MIDI and word durations as a reference for the SVS system to maintain pitch and rhythmic consistency.
\textbf{(2) Snippets Generation}:
We employ an in-house zero-shot SVS system to generate vocal snippets for each song, conditioned on the extracted MIDI and structural information from the previous stage. The synthesis is additionally conditioned on target speaker embeddings to ensure high-fidelity vocals that are both natural and timbre-consistent.
\textbf{(3) Instruction Design}:
We systematically enumerate various instruction formats, ranging from artist-specific queries to requests for particular song segments. We then leverage LLMs to paraphrase these templates into natural and colloquial questions. Finally, following the established pipeline for conversational data, all textual instructions are converted into query speech.



%% file: sections/4.exp.tex
\section{Experiments}

We evaluate \algname in role-playing, singing, and natural conversation benchmarks. In addition, training details, \algnamens's ASR and TTS results, ablation studies on speech tokenization and speaker injection are provided in Appendix.

\subsection{Role-playing}
\input{tables/role-play-simple}

We evaluate textual role-playing performance on the CharacterEval benchmark ~\citep{tu2024charactereval}. The assessment covers three dimensions: 
Character {\bf Consistency}, 
{\bf Conversation} Ability , 
and Role-playing {\bf Attractiveness}.
We construct character profiles from audiobook data, which are typically more concise and lack extensive background information compared to traditional role-playing corpora. An LLM is used to select conversations corresponding to specific evaluation dimensions, and role-playing–oriented models from the CharacterEval benchmark are employed to score performance along these dimensions.
Additionally, we evaluate speech performance using speaker similarity scores between generated speech and ground-truth responses.

The baselines comprise general-purpose LLM such as Baichuan2-7B ~\citep{yang2023baichuan} and Qwen3-8B ~\citep{yang2025qwen3}, chat agents such as 
Qwen-7B-Chat ~\citep{qwen7b}, GLM-9B-Chat~\citep{glm2024chatglm} and XVERSE-7B-Chat~\citep{XVERSE-7B}, role-playing specialized model such as Qwen2.5-7B-Instruct~\citep{qwen2025qwen25technicalreport}. All these baselines are text-only models and lack speech generation capability. We further include SLMs, such as Qwen2.5-Omni~\cite{qwen2025qwen25technicalreport} and Kimi-Audio~\cite{kimiteam2025kimiaudiotechnicalreport}, for comparison. Notably, none of the peer models support dynamic timbre control, which may limit the fairness of the comparison. The results are shown in Table~\ref{tab:role_play_simple}. The original Consistency, Conversation, and Attractiveness scores are measured on a 5-point scale and are linearly rescaled to a 100-point scale to align with speaker similarity scores.
We observe that our method achieves competitive text-level role-playing performance relative to conventional role-playing models. In particular, \algnamens-8B demonstrates strong performance in both conversational ability and attractiveness. Both \algnamens-4B and \algnamens-8B achieve speaker similarity scores of approximately $64\%$, indicating a reasonable alignment between synthesized timbre and the target role description. The higher speaker similarity compared to Qwen2.5-Omni and Kimi-Audio is primarily due to their lack of support for role-conditioned timbre control.
Detailed textual evaluations are shown in Appendix.

Additionally, at the speech level, we conduct subjective mean opinion score (MOS) evaluations to assess the quality of role-playing interactions. Five annotators rate each sample on a $5$-point scale. The evaluation framework measures performance along three dimensions: character matching score~\citep{dai2025deep} (\textbf{CMS}), which assesses the consistency between synthesized speech timbre and user-defined profiles; perceptual naturalness (\textbf{MOS-N}), which assesses acoustic fidelity; and response emotion (\textbf{MOS-E}), which assesses the appropriateness of the expressed emotion.
As shown in Table~\ref{tab:role_play_simple}, the results demonstrate strong MOS performance for role-playing. In particular, \algnamens-4B and \algnamens-8B achieve a CMS of approximately $3.5$, indicating good alignment between the generated voice and user expectations. It also attains high MOS-N and MOS-E scores, suggesting that the synthesized speech is both natural and emotionally appropriate. These results indicate that the proposed method effectively aligns both content and timbre with the target character profile. The scores of Qwen2.5-Omni and Kimi-Audio are included for reference as fixed-timbre SLMs.

\subsection{Singing}
For objective evaluation, we employ the latest FunASR model~\citep{an2025fun} to transcribe generated singing and compute the word error rate (\textbf{WER}) for assessing pronunciation accuracy. We further use \textbf{SingMOS}~\citep{tang2025singmos} and \textbf{Sheet-SSQA}~\citep{huang2024mos} to evaluate the perceptual quality of the generated singing.
For subjective evaluation, three annotators assess $100$ randomly sampled singing generation requests, from which MOS scores are computed. We adopt standard SVS metrics, including sound quality (\textbf{Qua.}), pronunciation clarity (\textbf{Clar.}), naturalness (\textbf{Nat.}), and expressiveness (\textbf{Expr.}).
\algname is compared with an SVS model conditioned on the corresponding MIDI files in both objective and subjective evaluations, and with Doubao soul singer in subjective evaluations.


\input{tables/song_table}


As shown in Table~\ref{tab:song_table}, \algname achieves singing quality--measured by SingMOS and Sheet-SSQA scores--comparable to specialized in-house SVS models. It attains a WER of approximately $0.2$, indicating reasonable singing accuracy. However, pronunciation clarity, audio quality, and expressiveness remain inferior to those of the SVS model, likely due to its reliance on MIDI guidance and the absence of song-specific training for our audio tokenizer. Overall, these results demonstrate the feasibility of integrating singing capabilities into E2E SLMs. Compared with Doubao Soul Singer, \algname achieves higher overall MOS across all four subjective dimensions, indicating superior singing voice quality.



\subsection{Natural Conversation Benchmark}
We evaluate the natural conversation ability of \algname on two spoken dialogue benchmark: C3 Benchmark~\citep{ma2025c3} and URO-Bench~\citep{yan2025uro}.

\input{tables/c3.tex}

C3 Benchmark is a bilingual benchmark for SLMs in evaluating their complex conversation capabilities. It decomposes conversational complexity into two dimensions: ambiguity and context dependency. Ambiguity is further evaluated through two subtests—phonological ambiguity and semantic ambiguity—while context dependency consists of three subtests: omission, coreference, and multi-turn interaction. The ambiguity and context-dependency scores are computed by averaging their respective subtests, and the overall score is obtained by averaging across all five subtests. We compare \algname against Step-Audio~\citep{huang2025step}, Qwen2.5-Omni~\citep{xu2025qwen2}, Kimi-Audio~\citep{ding2025kimi} and GLM-4-Voice~\citep{zeng2024glm}.
The results are shown in Table~\ref{tab:c3}. We observe that \algnamens-8B achieves the second-highest overall score on English, following Qwen2.5-Omni, and the highest on Chinese. \algnamens-8B performs better on resolving context-dependency related tasks but is weaker at handling ambiguity-related questions.


\input{tables/uro_ucr.tex}
We evaluate \algname on URO-Bench, which comprises three subtests: Understanding, Oral Conversation, and Reasoning. In addition to reporting subtest results, we compute the average score across all three. We compare \algname with Freeze-Omni, Qwen2.5-Omni, Kimi-Audio, GLM-4-Voice, and Chroma~\citep{Chen2026FlashLabsC1}. Notably, Chroma is a recent end-to-end SLM designed for voice cloning and natural conversation and has a model scale comparable to \algnamens-4B. The results are presented in Table~\ref{tab:uro_ucr}. We observe that \algnamens-8B ranks first in both English and Chinese among the compared models. \algnamens-4B lags approximately $7$ percentage points behind \algnamens-8B, ranking third in English, with an average score $1.15$ percentage points lower than Qwen2.5-Omni and $1.07$ percentage points lower than GLM-4-Voice and ranking second in Chinese, with a $2.85$ percentage point gap to GLM-4-Voice. Despite a similar model scale, \algnamens-4B outperforms Chroma by $11.44$ percentage points in English.

%% file: tables/role-play-simple.tex
\begin{table*}[t]
	\caption{Objective and subjective results for role-playing in text and speech. Text is evaluated on Character Consistency ({\bf CST}), Conversation Ability ({\bf CNV}), and Attractiveness ({\bf ATR}); speech is assessed by Speaker Similarity ({\bf SS}) against ground truth. Overall performance is averaged ({\bf Avg}). Subjective speech evaluation includes Character Matching ({\bf CMS}), Naturalness ({\bf MOS-N}), and Emotion ({\bf MOS-E}), all on a 5-point scale.
    }
	\label{tab:role_play_simple}
	\begin{center}
		\begin{adjustbox}{max width=0.9\textwidth}
			\begin{tabular}{lccccc|cccccc}
				\toprule
                \multirow{2}{*}{\bf Model} & \multicolumn{5}{c|}{\textbf{Objective}} & \multicolumn{3}{c}{\textbf{Subjective}} \\
                \cmidrule(lr){2-6} \cmidrule(lr){7-9} 
                &  CST &  CNV &  ATR &  SS & Avg. & CMS & MOS-N & MOS-E \\\midrule
                Baichuan2-7B     & 48.26 & 73.94 & 53.45 & - & - & - & - & -\\
                Qwen3-8B         & \underline{50.37} & 74.82 & 55.28 & - & - & - & - & -\\
                Qwen-7B-Chat     & 47.33 & 72.44 & 52.44 & - & - & - & - & -\\
                XVERSE-7B-Chat   & 48.47 & 74.07 & 53.60 & - & - & - & - & -\\
                GLM-9B-Chat      & 50.34 & \underline{75.42} & 55.20 & - & - & - & - & -\\
                Qwen2.5-7B-Instruct & 49.05 & 74.80 & 54.27 & - & - & - & - & -\\
                \midrule
                Qwen2.5-Omini    & 50.24 & 73.52 & 55.12 & 38.90 & 54.45 & 2.21 & 3.32 & 3.16\\
                KimiAudio        & {\bf50.48} & 75.30 & \underline{55.22} & 35.18 & 54.04 & 2.17 & 3.64 & 3.38\\
        
                \algnamens-4B    & 48.60 & 73.90 & 53.82 & \underline{64.07} & \underline{60.13} & \bf 3.54 & \bf 3.84 & \bf 4.08\\
                \algnamens-8B    & 50.34 & \bf 75.48 & \bf 55.31 & \bf 64.70 & \bf 61.45  & \underline{3.45} & \underline{3.74} & \underline{3.99}\\

				\bottomrule
			\end{tabular}
		\end{adjustbox}
	\end{center}
\end{table*}

%% file: tables/song_table.tex
\begin{table*}[t]
    \caption{Objective and subjective evaluation results on singing. Objective metrics include SingMOS, Sheet-SSQA, and WER, where SingMOS and Sheet-SSQA scores are for singing voice perceptual quality, and WER is for pronunciation accuracy. Subjective metrics include sound quality (Qua.), pronunciation clarity (Clar.), naturalness (Nat.), and expressiveness (Expr.). All scores are reported on a 5-point scale.}
    \label{tab:song_table}
    \begin{center}
        \begin{adjustbox}{max width=0.95\textwidth}
            \begin{tabular}{l ccc @{\hskip 1cm} cccc}
                \toprule
                \multirow{2}{*}{\textbf{Model}} & \multicolumn{3}{c}{\textbf{Objective Metrics}} & \multicolumn{4}{c}{\textbf{Subjective MOS}} \\
                \cmidrule(lr){2-4} \cmidrule(lr){5-8}
                 & SingMOS~$\uparrow$  & Sheet-SSQA~$\uparrow$ & WER~$\downarrow$ & Qua. & Clar. & Nat. & Expr. \\
                \midrule
                In-house SVS     & 4.148  & 2.534 & 0.118 & 4.45  & 4.64  & 4.46  & 4.55 \\\midrule
                Doubao Soul Singer   & - & - & - & 3.10  &3.89  &3.37  &3.19 \\ 
                \algnamens-4B    & 4.040  & 3.198 & 0.211 & 3.25  & 4.11  & 3.71  & 3.21 \\
                \algnamens-8B    & 4.043  & 3.222 & 0.266 & 3.31  & 4.05  & 3.49  & 3.22 \\
                \bottomrule
            \end{tabular}
        \end{adjustbox}
    \end{center} 
\end{table*}

%% file: tables/c3.tex
\begin{table*}[t]
	\caption{Results on C3 Benchmarks. The best results are highlighted in {\bf bold}, and the second-best are  \underline{underlined}.}
	\label{tab:c3}
	\begin{center}
		\begin{adjustbox}{max width=0.99\textwidth}
			\begin{tabular}{lccccccccc}
				\toprule
                EN & \#Param & Phono & Semantic & Ambiguity & Omission & Coref. & MultiTurn & ContextDep & Overall \\

				\midrule
                Step-Audio& 130B & 29.31 & 21.57 & 25.44 & 10.78 & 57.31 & 41.18 & 36.43 & 32.03 \\
                Qwen2.5-Omni& 7B & \bf 48.28 & \bf 32.35 & \bf 40.31& 15.2 & 68.15 & \bf 95.59 & \bf59.64& \bf 51.91 \\
                Kimi-Audio& 7B & \underline{46.55} & \underline{29.41} & \underline{37.98} & 10.29 & \bf 87.41 & -- &48.85&43.42 \\
                GLM-4-Voice& 9B & 27.59 & 15.69 & 21.64 & 6.37 & \underline{68.98} & 58.82 & 44.73 & 35.49 \\
                Freeze-Omni& 7B & 8.62 &  11.76 & 10.19 & 6.86 &  47.22 & 44.12 & 32.73 & 23.72 \\



				\midrule
                \algnamens-4B& 4B & 20.69  & 4.71 & 12.70  & \bf 20.39 & 64.52 & 74.12 & \underline{53.01} & 36.89 \\
                \algnamens-8B & 8B & 41.38  & 21.57 & 31.47  & \underline{17.84} & 64.15 & \underline{75.29} & 52.43 & \underline{44.05}  \\

				
				
                \midrule\midrule
                ZH & \#Param & Phono & Semantic & Ambiguity & Omission & Coref. & MultiTurn & ContextDep & Overall \\
                \midrule
                Step-Audio& 130B & 18.92 & 2.54 &  10.73 & 5.71 &  16.67 & 10.53 & 10.97 & 10.87 \\
                Qwen2.5-Omni& 7B  & \bf 27.03& \underline{6.78}& \underline{16.9}& 27.86& 55.83& \bf 82.89& 55.53 & \underline{40.08} \\
				Kimi-Audio& 7B & 20.27& 4.24& 12.25& 29.29& 40& --& 34.64& 23.45 \\
                GLM-4-Voice& 9B & \underline{22.97} & 5.08 &  14.03 & 17.86 & 50.83 & 7.89 &  25.53 & 20.93 \\
                Freeze-Omni& 7B & 16.22 & 1.69 &  8.96 &  4.29 &  10.83 & 11.84 & 8.99 &  8.97 \\
                \midrule
                \algnamens-4B& 4B & 9.19  & 2.37 & 5.78  & \underline{61.14} & \underline{60} & 47.37 & \underline{56.17} & 36.01 \\
                \algnamens-8B& 8B & 13.51  & \bf 22.03 & \bf{17.77}  & \bf 74.29 & \bf{70} & \underline{73.68} & \bf 72.66 & \bf 50.70 \\
				\bottomrule
			\end{tabular}
		\end{adjustbox}
		
	\end{center}
\end{table*}

%% file: tables/uro_ucr.tex
\begin{table*}[t]
	\caption{Results on URO Benchmarks. ``U.'', ``C.'', ``R.'', and ``Avg.'' denote ``Understanding'', ``Conversation'', ``Reasoning'', and ``Average'' respectively. 
    }
	\label{tab:uro_ucr}
    \small
	\begin{center}
		\begin{adjustbox}{max width=0.8\textwidth}
			\begin{tabular}{lccccccccccc}
				\toprule
				\multirow{2}{*}{{Model}} & \multirow{2}{*}{{\#Param}}  &  \multicolumn{4}{c}{\textbf{EN}} & \multicolumn{4}{c}{\textbf{ZH}}\\
                \cmidrule(lr){3-6}\cmidrule(lr){7-10}
                 & &U. & C. & R. & Avg. &U. & C. & R. & Avg. \\
				\midrule
                Freeze-Omni & 7B & 58.68 & 52.24 & 37.52 & 49.48 & 28.15 & 68.05 & 21.27 & 39.16 \\
                Qwen2.5-Omni& 7B & 66.29 & \bf{76.16} & \bf 69.62 & \underline{70.69} & 46.44 & \underline{75.06} & \bf 67.55 & 63.02 \\
                KimiAudio& 7B & \underline{83.36} & 60.36 & 42.31 & 62.01 & 75.86 & 75.86 & 59.69 & 70.47\\
                GLM-4-Voice& 9B & 82.16 & 74.20 & 55.46 & 70.61 & \underline{88.57} & \bf 83.32 & 45.09 & \underline{72.33} \\
                Chroma& 4B & 60.59 & 62.26 & 51.46 & 58.10 & --&--&--&-- \\
				\midrule
				\algnamens-4B& 4B & 82.31  & 70.21  & 56.10  & 69.54 & 82.41  & 69.60  & 56.43  & 69.48 \\
                \algnamens-8B& 8B & \bf89.91 & \underline{73.64} & \underline{66.32} & \bf{76.62} & \bf{89.76}  & 74.50  & \underline{64.55}  & \bf 76.27  \\

				\bottomrule
			\end{tabular}
		\end{adjustbox}
	\end{center}
\end{table*}

%% file: sections/5.concl.tex
\section{Conclusion}
In this work, we present \algname, the first end-to-end spoken language model that is capable of not only natural conversations but also expressive speech generation, including role-playing and singing.
\algname adopts a novel hybrid text-speech modeling approach that enables the native end-to-end learning of rich paralinguistic features. Both subjective and objective evaluations demonstrate that \algname achieves state-of-the-art performance on spoken dialog benchmarks, while exhibiting strong singing and role-playing abilities. 
The role-playing and singing capabilities introduced in this work represent early exploratory efforts toward broader expressive speech generation. 
We hope that this study can provide a useful starting point for future research.

%% file: sections/append.tex
\section{Pretraining Data}
\textbf{TTS Data}\quad We leverage a massive dataset totaling approximately 867K hours for Text-to-Speech (TTS) training. We utilize a TTS data pipeline (audio denoising, speaker diarization, and ASR) to curate 762K hours of high-quality corpus from diverse sources, including audiobooks, podcasts, children's stories, and traditional Chinese performing arts. Concurrently, we integrate 105K hours of open-source TTS data, mainly consisting of the WenetSpeech4TTS~\citep{ma2024wenetspeech4tts}, LibriTTS~\citep{zen2019libritts}, GLOBEv2~\citep{wang2024globehighqualityenglishcorpus}, and Emilia~\citep{he2024emilia} datasets. This combination of large-scale proprietary recordings and diverse public corpora ensures superior prosodic richness and multi-speaker modeling.

\textbf{ASR Data}\quad We aggregate approximately 100K hours of open-source Automatic Speech Recognition (ASR) data, including WenetSpeech~\citep{zhang2022wenetspeech}, Librispeech~\citep{zen2019libritts}, Mls ~\citep{pratap2020mls}, Common Voice ~\citep{ardila2020common}, SLR68, GigaSpeech~\citep{chen2021gigaspeech}, People's Speech~\citep{galvez2021people}, VoxPopuli~\citep{wang2021voxpopuli}, and the AISHELL series~\citep{bu2017aishell, du2018aishell, shi2020aishell, fu2021aishell}. 

\textbf{SQA Data}\quad
Speech-to-text understanding takes spoken questions as input and generates textual responses as output. This task is ignored in native SLMs~\citep{xie2024mini, gao2025lucy, long2025vita}, but it has been shown to significantly preserve the intelligence of LLMs in recent modular SLMs~\citep{chen2025minmo, xu2025qwen3omnitechnicalreport}. As a result, we develop a Speech Question Answering (SQA) dataset selected from in-house QA text pairs. The SQA dataset covers approximately 18K hours of question speech, including general knowledge, commonsense reasoning, and reading comprehension.

\textbf{Text Data}\quad
Following VITA-Audio~\citep{Long2025VITAAudioFI}, we collect open-source text corpora of $11$B tokens covering conversations on knowledge answering, understanding, reasoning, and long-context text.

\section{Training Details}

We train \algname using a standard align-pretrain-SFT training pipeline. 
Stage $0$ is for modality alignment, where we freeze the audio encoder and the backone LLM and train only the MLP-based audio adapter $10\%$ of pretraining data to align the audio encoder's output with the LLM's input space. The adapter is trained $500$ steps with an effective batch size of $1.28$M and a learning rate of $1e$-$3$.
Stage $1$ is for pretraining, where we train both the audio adapter and the LLM on our full open-source and in-house ASR, TTS, SQA, and text pretraining data as described in Sec~\ref{sec:pretraining_data}. The model is trained $8$K steps with an effective batch size of $2.56$M tokens and a learning rate of $6e$-$5$.
Stage $2$ is for supervised finetuning, where we train both the audio adapter and the LLM on mainly spoken dialogue data, including natural conversation, role-playing, and singing data, as well as a small portion of pretraining data from the previous stage.
We train $8$K steps for Stage $2$ with an effective batch size of $1.28$M tokens and a learning rate of $6e$-$5$. 
Following VITA-Audio~\citep{Long2025VITAAudioFI}, all training data are packed into sequences with a fixed length of $10$K tokens to maximize GPU utility. 

\section{ASR and TTS}
\input{tables/asr.tex}

\textbf{ASR}\quad
We evaluate the ASR performance of \algname on WenetSpeech~\citep{zhang2022wenetspeech} and 
AIShell~\citep{bu2017aishell} datasets for Chinese and LibriSpeech~\citep{panayotov2015librispeech} for English. We compute character error rates(CER) and word error rates (WER) for Chinese and English, respectively. The results are shown in Table~\ref{tab:asr}. The performance of Qwen3-Omni~\citep{xu2025qwen3omnitechnicalreport} with 30B parameters is provided as a top-line reference.
We observe that \algnamens-8B achieves lower error rates than \algnamens-4B, possibly due to stronger language capabilities, and \algnamens-4B and \algnamens-8B have lower error rates than the GLM-4-Voice~\citep{zeng2024glm} and Freeze-Omni~\citep{wang2024freeze} models in the three datasets and achieve a performance comparable to that of Qwen2.5-Omni~\cite{qwen2025qwen25technicalreport}. 

\textbf{TTS}\quad
\input{tables/tts.tex}

We evaluate the TTS performance of \algname on the Seed-TTS~\citep{Anastassiou2024SeedTTSAF} benchmark. The generated English speech is transcribed into text using Whisper-Large-V3~\citep{radford2023robust} and Chinese speech using Paraformer~\citep{gao2022paraformer}. The CER and WER are computed by comparing the resulting transcription and the ground-truth text for Chinese and English, respectively. 
We observe that \algnamens-8B and \algnamens-4B perform comparably, with the 4B variant slightly better in English and the 8B variant slightly better in Chinese. In particular, \algnamens-8B produces speech more faithfully aligned with the input text than GLM-4-Voice and Qwen2.5-Omni in Chinese, achieving state-of-the-art performance comparable to, and occasionally surpassing, specialized TTS models such as Seed-TTS~\citep{Anastassiou2024SeedTTSAF} and CosyVoice2~\citep{du2024cosyvoice2}.

\section{Detailed Role-playing Results}
\input{tables/role-play-obj-detail}
Detailed role-playing results are shown in Table~\ref{tab:role_play_obj_detail}.

\section{Multi- v.s. Single-Codebook Speech Tokenizers}
\input{tables/dtw.tex}
\textbf{Pitch Contour Comparison}\quad 
In our preliminary experiments, we compare the multi-codebook XY-Tokenizer with the single-codebook CosyVoice2 Tokenizer~\citep{du2024cosyvoice2} and GLM-4-Voice Tokenizer~\citep{zeng2024glm} for singing voice reconstruction. XY-Tokenizer operates at 100 Hz, whereas CosyVoice2 and GLM-4-Voice use lower token rates of 25 Hz and 12.5 Hz, respectively. The CosyVoice2 decoder additionally requires speaker embeddings to recover speaker identity, while GLM-4-Voice neither requires nor preserves speaker identity. We randomly sample 100 songs from the singing data and reconstruct them using each tokenizer. The normalized pitch contours of the reconstructed audio are extracted and compared with those of the original recordings. Mean squared error (MSE) and dynamic time warping (DTW) distances are computed for quantitative evaluation. The results, shown in Table~\ref{tab:dtw}, indicate that XY-Tokenizer achieves the best reconstruction with the lowest MSE and DTW distance. Consistent with this, qualitative listening indicates that the XY-Tokenizer preserves the original melody, while the other two methods miss most melodic variations.

\input{tables/utmos.tex}
\textbf{UTMOS Comparison}\quad
We then compare \algname using XY-Tokenizer~\citep{gong2025xy} with VITA-Audio using the GLM-4-Voice Tokenizer~\citep{zeng2024glm}. Speech responses are generated in the URO Benchmark~\citep{yan2025uro}, and UTMOS~\citep{baba2024utmosv2} scores—a reference-free metric that assesses speech naturalness—are computed. The results are summarized in Table~\ref{tab:utmos}. We observe that \algnamens-4B with XY-Tokenizer achieves higher UTMOS scores in both English and Chinese, despite its smaller model scale.

\section{Ablation on Speaker Pretraining}
\input{tables/role-play-nospk}
We conduct an ablation study to evaluate the effect of speaker pretraining. Specifically, in Stage 1 (pretraining), we remove both agent and user embedding information from the input and then train Stage 2 from the Stage 1 checkpoint, resulting in \algnamens-8B-\textsc{nospk}. We compare its performance on role-playing responses with that of \algnamens-8B. The results are presented in Table~\ref{tab:role_play_nospk}.
We observe that \algnamens-8B-\textsc{nospk} underperforms \algnamens-8B in both subjective and objective evaluations, indicating that explicitly injecting speaker information during pretraining effectively decouples content and timbre. This facilitates better modeling of both text and speech, thus improving role-playing performance.

%% file: tables/asr.tex
\begin{table*}[ht]
	\caption{Results on Automatic Speech Recognition~(ASR) Benchmarks. The best results are highlighted in
{\bf bold}, and the second-best are \underline{underlined}.}\label{tab:asr}
	\begin{center}
		\begin{adjustbox}{max width=0.99\textwidth}
			\begin{tabular}{lccccccc}
				\toprule
				\multirow{2}{*}{{Model}} & \multirow{2}{*}{{\#Param}} & \multicolumn{2}{c}{{WenetSpeech}} & {AIShell} & \multicolumn{2}{c}{{LibriSpeech}} \\\cmidrule(lr){3-4}\cmidrule(lr){5-5}\cmidrule(lr){6-7}
				&& test\_net~$\downarrow$ & test\_meeting~$\downarrow$ & test~$\downarrow$ & test-clean~$\downarrow$ & test-other~$\downarrow$ \\
				\midrule
                Qwen3-Omni & 30B & 4.69 &  5.89 & -    &  1.22 & 2.48 \\	\midrule
				GLM-4-Voice & 9B & - & - & 2.46 &2.82 & 7.66 \\
				Freeze-Omni & 7B & 11.80 & 13.46 & 2.48 & 3.82 & 9.79 \\
                VITA-Audio & 7B   & 6.68 & \underline{6.59} & \underline{1.51} & 1.91 & 4.29 \\
                Qwen2.5-Omni & 7B  & \bf{6.04} &  7.71 & \bf{1.13} & \underline{1.80} & \bf{3.40} \\


				\midrule
				\algnamens-4B & 4B & 6.75 & 7.08 & 2.35 & 2.25 & 5.47 \\
				\algnamens-8B & 8B & \underline{6.42} & \bf{6.56} & 1.64 & \bf{1.75} & \underline{4.23} \\

				

				\bottomrule

			\end{tabular}
		\end{adjustbox}
		
	\end{center}
\end{table*}

%% file: tables/tts.tex
\begin{table*}[ht]
	\caption{Results on Automatic Speech Recognition~(ASR) Benchmarks. The best results are highlighted in
{\bf bold}, and the second-best are \underline{underlined}.
    }
	\label{tab:tts}
	\begin{center}
		\begin{adjustbox}{max width=0.99\textwidth}
			\begin{tabular}{lccccc}
				\toprule
				\multirow{2}{*}{{Model}} & \multirow{2}{*}{{\#Param}} & \multicolumn{2}{c}{{Seed-TTS}}\\\cmidrule(lr){3-4}
				&& zh (CER)~$\downarrow$ & en (WER)~$\downarrow$ \\
				\midrule
                Qwen3-Omni & 30B & 1.07 & 1.39 \\\midrule
                Seed-TTS & - & \underline{1.12} & 2.25 \\
                CosyVoice2 & 0.5B & 1.45 & 2.57 \\
                GLM-4-Voice & 9B & 2.99 & \bf 2.10 \\
                Qwen2.5-Omni & 7B & 1.42 & 2.33 \\



				\midrule
				\algnamens-4B & 4B & 1.19 & \underline{2.20}  \\
				\algnamens-8B & 8B & \bf 1.01 & 2.29 \\

				

				\bottomrule

			\end{tabular}
		\end{adjustbox}
		
	\end{center}
\end{table*}

%% file: tables/role-play-obj-detail.tex
\begin{table*}[t]
    \caption{Detailed objective evaluation results on generated text response for role-playing. The text responses are evaluated across character consistency ({\bf Consistency}), conversational ability (\textbf{Conversation}), and role-playing attractiveness (\textbf{Attractiveness}). 
    Character Consistency is evaluated on Knowledge-Exposure (KE), Knowledge-Accuracy (KA), Knowledge-Hallucination (KH), Persona-Behavior (PB), and Persona-Utterance (PU);
    Conversational ability is evaluated on Fluency (Flu.), Coherency (Coh.), and Consistency (Cons.);
    Attractiveness is evaluated on Human-Like (HL), Communication Skills (CS), Expression Diversity (ED), and Empathy (Emp.).}
    \label{tab:role_play_obj_detail}
    \centering
    \small
    \setlength{\tabcolsep}{2pt}

    \begin{tabular*}{\textwidth}{@{\extracolsep{\fill}} lcccccc|cccc|cccccc }
        \toprule
        \multirow{2}{*}{\textbf{Model}} & \multicolumn{6}{c}{\textbf{Consistency}} & \multicolumn{4}{c}{\textbf{Conversation}} & \multicolumn{5}{c}{\textbf{Attractiveness}}\\
        \cmidrule(lr){2-7} \cmidrule(lr){8-11} \cmidrule(lr){12-16} 
         & KE & KA & KH & PB & PU & \multicolumn{1}{c}{Avg.} & Flu. & Coh. & Cons. & \multicolumn{1}{c}{Avg.} & HL & CS & ED & Emp & \multicolumn{1}{c}{Avg.} \\
        \midrule
        Baichuan2-7B     & 1.83 & 3.03 & 2.70 & 1.33 & 2.59 & 2.41 & 3.53 & 3.88 & 3.69 & 3.70 & 3.10 & 2.97 & 1.36 & 3.26 & 2.67 \\
        Qwen3-8B         & 1.83 & 3.12 & 2.74 & \textbf{1.58} & 2.64 & \underline{2.52} & 3.59 & 3.91 & 3.73 & 3.74 & 3.21 & 2.99 & \textbf{1.53} & 3.33 & \underline{2.76} \\
        ChatGLM3-6B      & \underline{1.87} & 2.97 & 2.57 & 1.29 & 2.44 & 2.32 & 3.32 & 3.70 & 3.39 & 3.47 & 2.84 & 2.93 & 1.33 & 3.14 & 2.56 \\
        Qwen-7B-Chat     & 1.84 & 2.99 & 2.62 & 1.35 & 2.50 & 2.37 & 3.46 & 3.83 & 3.58 & 3.62 & 2.94 & 2.96 & 1.38 & 3.21 & 2.62 \\
        XVERSE-7B-Chat   & 1.83 & 3.05 & 2.72 & 1.32 & 2.60 & 2.42 & 3.53 & 3.88 & 3.70 & 3.70 & 3.11 & 2.98 & 1.35 & 3.28 & 2.68 \\ 
        GLM-9B-Chat      & 1.75 & 3.10 & 2.73 & \underline{1.55} & \underline{2.69} & 2.52 & 3.61 & 3.91 & \textbf{3.80} & \underline{3.77} & \underline{3.34} & 2.87 & \underline{1.52} & 3.32 & 2.76 \\
        Qwen2.5-7B-Instruct & 1.86 & 3.08 & \underline{2.75} & 1.36 & 2.63 & 2.45 & 3.57 & \textbf{3.92} & 3.74 & 3.74 & 3.13 & 3.03 & 1.38 & 3.32 & 2.71 \\
        \midrule
        Qwen2.5-Omini    & \textbf{1.88} & \underline{3.38} & 2.74 & 1.22 & \textbf{2.71} & 2.51 & 3.56 & 3.81 & 3.66 & 3.68 & 3.24 & \textbf{3.06} & 1.26 & \underline{3.46} & 2.76 \\
        Kimi-Audio        & \underline{1.87} & \textbf{3.41} & \textbf{2.75} & 1.24 & 2.69 & \textbf{2.52} & \textbf{3.62} & 3.89 & 3.79 & 3.77 & 3.23 & \underline{3.04} & 1.26 & \textbf{3.50} & 2.76 \\

        \midrule
        \algnamens-4B    & 1.81 & 3.08 & 2.71 & 1.35 & 2.61 & 2.44 & 3.54 & 3.87 & 3.68 & 3.70 & 3.15 & 2.98 & 1.35 & 3.30 & 2.69 \\
        \algnamens-8B    & 1.75 & 3.12 & 2.73 & 1.53 & 2.69 & 2.52 & \underline{3.62} & \underline{3.91} & \underline{3.79} & \textbf{3.77} & \textbf{3.35} & 2.88 & 1.51 & 3.32 & \textbf{2.77} \\
        \bottomrule
    \end{tabular*}

\end{table*}

%% file: tables/dtw.tex
\begin{table*}[t]
	\caption{MSE and DTW distance for XY-Tokenizer, GLM-4-Voice Tokenizer and Cosyvoice2 Tokenizer.}
	\label{tab:dtw}
	\begin{center}
		\begin{adjustbox}{max width=0.99\textwidth}
			\begin{tabular}{lccccccccccc}
				\toprule
                Model  & MSE$\downarrow$ & DTW$\downarrow$ \\\midrule
                XY-Tokenizer & \bf0.213 & \bf9.162 \\
                CosyVoice2 & 0.374 & 13.063 \\
                GLM-4-Voice & 0.509 & 15.460 \\

				\bottomrule
			\end{tabular}
		\end{adjustbox}
	\end{center}
\end{table*}

%% file: tables/utmos.tex
\begin{table*}[ht]
	\caption{UTMOS Results for English and Chinese on the URO Benchmark.}
	\label{tab:utmos}
	\begin{center}
		\begin{adjustbox}{max width=0.99\textwidth}
			\begin{tabular}{lccccccccccc}
				\toprule
                Model & \#Param & Tokenizer & EN$\uparrow$ & ZH$\uparrow$ \\\midrule
                VITA-Audio & 7B & GLM-4-Voice & 4.10 & 2.61 \\
                \algnamens-4B & 4B & XY-Tokenizer & \bf 4.34 & \bf 3.56 \\

				\bottomrule
			\end{tabular}
		\end{adjustbox}
	\end{center}
\end{table*}

%% file: tables/role-play-nospk.tex
\begin{table*}[ht]
	\caption{Objective evaluation results for role-playing in both text and speech. Text responses are assessed in terms of character consistency ({\bf Consistency}), conversational ability ({\bf Conversation}), and role-playing attractiveness ({\bf Attractiveness}), while speech responses are evaluated by speaker similarity with respect to ground-truth speech. The overall performance is reported as the average score ({\bf Avg.}) across these four dimensions. The best results are highlighted in \textbf{bold}, and the second-best are \underline{underlined}.}
	\label{tab:role_play_nospk}
	\begin{center}
		\begin{adjustbox}{max width=0.9\textwidth}
			\begin{tabular}{lccccccccccc}
				\toprule
                \multirow{2}{*}{\bf Model} & \multicolumn{3}{c}{\textbf{Text}} & \multicolumn{1}{c}{\textbf{Speech}}  \\
                \cmidrule(lr){2-4} \cmidrule(lr){5-5} 
                &  Consistency &  Conversation &  Attractiveness &  Speaker Similarity \\\midrule
                \algnamens-8B-\textsc{nospk}    & 48.6 & 73.11 & 52.81 & 64.02 \\
                \algnamens-8B    & \bf 50.34 & \bf 75.48 & \bf 55.31 & \bf 64.70 \\
				\bottomrule
			\end{tabular}
		\end{adjustbox}
	\end{center}
\end{table*}

%% file: reference.bib
@article{liu2024deepseek,
  title={Deepseek-v3 technical report},
  author={Liu, Aixin and Feng, Bei and Xue, Bing and Wang, Bingxuan and Wu, Bochao and Lu, Chengda and Zhao, Chenggang and Deng, Chengqi and Zhang, Chenyu and Ruan, Chong and others},
  journal={arXiv preprint arXiv:2412.19437},
  year={2024}
}

@inproceedings{baba2024utmosv2,
  title     = {The T05 System for The {V}oice{MOS} {C}hallenge 2024: Transfer Learning from Deep Image Classifier to Naturalness {MOS} Prediction of High-Quality Synthetic Speech},
  author    = {Baba, Kaito and Nakata, Wataru and Saito, Yuki and Saruwatari, Hiroshi},
  booktitle = {IEEE Spoken Language Technology Workshop (SLT)},
  year      = {2024},
  pages={818--824},
  doi={10.1109/SLT61566.2024.10832315},
}

@inproceedings{Chen2026FlashLabsC1,
  title={FlashLabs Chroma 1.0: A Real-Time End-to-End Spoken Dialogue Model with Personalized Voice Cloning},
  author={Tanyu Chen and Tairan Chen and Kai Shen and Zhenghua Bao and Zhihui Zhang and Man Yuan and Yi Shi},
  year={2026},
  url={https://api.semanticscholar.org/CorpusID:284860926}
}

@article{Anastassiou2024SeedTTSAF,
  title={Seed-TTS: A Family of High-Quality Versatile Speech Generation Models},
  author={Philip Anastassiou and Jiawei Chen and Jitong Chen and Yuanzhe Chen and Zhuo Chen and Ziyi Chen and Jian Cong and Lelai Deng and Chuang Ding and Lu Gao and Mingqing Gong and Peisong Huang and Qingqing Huang and Zhiying Huang and Yuanyuan Huo and Dongya Jia and Chumin Li and Feiya Li and Hui Li and Jiaxin Li and Xiaoyang Li and Xingxing Li and Lin Liu and Shouda Liu and Sichao Liu and Xudong Liu and Yuchen Liu and Zhengxi Liu and Lu Lu and Junjie Pan and Xin Wang and Yuping Wang and Yuxuan Wang and Zhengnan Wei and Jian Wu and Chao Yao and Yifeng Yang and Yuan-Qiu-Qiang Yi and Junteng Zhang and Qidi Zhang and Shuo Zhang and WenJie Zhang and Yang Zhang and Zilin Zhao and Dejian Zhong and Xiaobin Zhuang},
  journal={ArXiv},
  year={2024},
  volume={abs/2406.02430},
  url={https://api.semanticscholar.org/CorpusID:270226353}
}

@misc{SileroVAD,
  author = {Silero-Team},
  title = {Silero VAD: pre-trained enterprise-grade Voice Activity Detector (VAD), Number Detector and Language Classifier},
  year = {2024},
  publisher = {GitHub},
  journal = {GitHub repository},
  howpublished = {\url{https://github.com/snakers4/silero-vad}},
  commit = {insert_some_commit_here},
  email = {hello@silero.ai}
}

@article{du2024cosyvoice2,
  title={Cosyvoice 2: Scalable streaming speech synthesis with large language models},
  author={Du, Zhihao and Wang, Yuxuan and Chen, Qian and Shi, Xian and Lv, Xiang and Zhao, Tianyu and Gao, Zhifu and Yang, Yexin and Gao, Changfeng and Wang, Hui and others},
  journal={arXiv preprint arXiv:2412.10117},
  year={2024}
}

@article{wang2025spark,
  title={Spark-tts: An efficient llm-based text-to-speech model with single-stream decoupled speech tokens},
  author={Wang, Xinsheng and Jiang, Mingqi and Ma, Ziyang and Zhang, Ziyu and Liu, Songxiang and Li, Linqin and Liang, Zheng and Zheng, Qixi and Wang, Rui and Feng, Xiaoqin and others},
  journal={arXiv preprint arXiv:2503.01710},
  year={2025}
}

@article{hwang2025hiddensinger,
  title={HiddenSinger: High-quality singing voice synthesis via neural audio codec and latent diffusion models},
  author={Hwang, Ji-Sang and Lee, Sang-Hoon and Lee, Seong-Whan},
  journal={Neural Networks},
  volume={181},
  pages={106762},
  year={2025},
  publisher={Elsevier}
}

@article{wu2024toksing,
  title={Toksing: Singing voice synthesis based on discrete tokens},
  author={Wu, Yuning and Shi, Jiatong and Tang, Yuxun and Yang, Shan and Jin, Qin and others},
  journal={arXiv preprint arXiv:2406.08416},
  year={2024}
}

@inproceedings{pan2025synthetic,
  title={Synthetic Singers: A Review of Deep-Learning-based Singing Voice Synthesis Approaches},
  author={Pan, Changhao and Yao, Dongyu and Zhang, Yu and Guo, Wenxiang and Lu, Jingyu and Zhu, Zhiyuan and Zhao, Zhou},
  booktitle={Proceedings of the 14th International Joint Conference on Natural Language Processing and the 4th Conference of the Asia-Pacific Chapter of the Association for Computational Linguistics},
  pages={396--416},
  year={2025}
}

@inproceedings{kim2021conditional,
  title={Conditional variational autoencoder with adversarial learning for end-to-end text-to-speech},
  author={Kim, Jaehyeon and Kong, Jungil and Son, Juhee},
  booktitle={International Conference on Machine Learning},
  pages={5530--5540},
  year={2021},
  organization={PMLR}
}

@inproceedings{zhang2022visinger,
  title={Visinger: Variational inference with adversarial learning for end-to-end singing voice synthesis},
  author={Zhang, Yongmao and Cong, Jian and Xue, Heyang and Xie, Lei and Zhu, Pengcheng and Bi, Mengxiao},
  booktitle={ICASSP 2022-2022 IEEE International Conference on Acoustics, Speech and Signal Processing (ICASSP)},
  pages={7237--7241},
  year={2022},
  organization={IEEE}
}

@misc{TEN_Turn_Detection,
author = {TEN-Team},
title = {TEN Turn Detection: Turn detection for full-duplex dialogue communication

},
year = {2025},
url = {https://github.com/TEN-framework/ten-turn-detection},
}

@article{huang2025step,
  title={Step-Audio-AQAA: a Fully End-to-End Expressive Large Audio Language Model},
  author={Huang, Ailin and Li, Bingxin and Wang, Bruce and Wu, Boyong and Yan, Chao and Feng, Chengli and Wang, Heng and Zhou, Hongyu and Wang, Hongyuan and Li, Jingbei and others},
  journal={arXiv preprint arXiv:2506.08967},
  year={2025}
}

@article{chen2024persona,
  title={From persona to personalization: A survey on role-playing language agents},
  author={Chen, Jiangjie and Wang, Xintao and Xu, Rui and Yuan, Siyu and Zhang, Yikai and Shi, Wei and Xie, Jian and Li, Shuang and Yang, Ruihan and Zhu, Tinghui and others},
  journal={arXiv preprint arXiv:2404.18231},
  year={2024}
}

@article{gao2022paraformer,
  title={Paraformer: Fast and accurate parallel transformer for non-autoregressive end-to-end speech recognition},
  author={Gao, Zhifu and Zhang, Shiliang and McLoughlin, Ian and Yan, Zhijie},
  journal={arXiv preprint arXiv:2206.08317},
  year={2022}
}

@article{gong2025xy,
  title={XY-Tokenizer: Mitigating the Semantic-Acoustic Conflict in Low-Bitrate Speech Codecs},
  author={Gong, Yitian and Jin, Luozhijie and Deng, Ruifan and Zhang, Dong and Zhang, Xin and Cheng, Qinyuan and Fei, Zhaoye and Li, Shimin and Qiu, Xipeng},
  journal={arXiv preprint arXiv:2506.23325},
  year={2025}
}

@article{an2024funaudiollm,
  title={Funaudiollm: Voice understanding and generation foundation models for natural interaction between humans and llms},
  author={An, Keyu and Chen, Qian and Deng, Chong and Du, Zhihao and Gao, Changfeng and Gao, Zhifu and Gu, Yue and He, Ting and Hu, Hangrui and Hu, Kai and others},
  journal={arXiv preprint arXiv:2407.04051},
  year={2024}
}

@article{lu2025youtu,
  title={Youtu-LLM: Unlocking the Native Agentic Potential for Lightweight Large Language Models},
  author={Lu, Junru and Qin, Jiarui and Qiao, Lingfeng and Li, Yinghui and Dai, Xinyi and Ke, Bo and He, Jianfeng and Qiao, Ruizhi and Yin, Di and Sun, Xing and others},
  journal={arXiv preprint arXiv:2512.24618},
  year={2025}
}

@inproceedings{ma2025c3,
  title={C3: A bilingual benchmark for spoken dialogue models exploring challenges in complex conversations},
  author={Ma, Chengqian and Tao, Wei and Guo, Steven Y},
  booktitle={Proceedings of the 2025 Conference on Empirical Methods in Natural Language Processing},
  pages={22789--22807},
  year={2025}
}

@article{yan2025uro,
  title={Uro-bench: A comprehensive benchmark for end-to-end spoken dialogue models},
  author={Yan, Ruiqi and Li, Xiquan and Chen, Wenxi and Niu, Zhikang and Yang, Chen and Ma, Ziyang and Yu, Kai and Chen, Xie},
  journal={arXiv preprint arXiv:2502.17810},
  year={2025}
}

@misc{pan2026syntheticsingersreviewdeeplearningbased,
      title={Synthetic Singers: A Review of Deep-Learning-based Singing Voice Synthesis Approaches}, 
      author={Changhao Pan and Dongyu Yao and Yu Zhang and Wenxiang Guo and Jingyu Lu and Zhiyuan Zhu and Zhou Zhao},
      year={2026},
      eprint={2601.13910},
      archivePrefix={arXiv},
      primaryClass={eess.AS},
      url={https://arxiv.org/abs/2601.13910}, 
}

@misc{zhang2025omnicharacterimmersiveroleplayingagents,
      title={OmniCharacter: Towards Immersive Role-Playing Agents with Seamless Speech-Language Personality Interaction}, 
      author={Haonan Zhang and Run Luo and Xiong Liu and Yuchuan Wu and Ting-En Lin and Pengpeng Zeng and Qiang Qu and Feiteng Fang and Min Yang and Lianli Gao and Jingkuan Song and Fei Huang and Yongbin Li},
      year={2025},
      eprint={2505.20277},
      archivePrefix={arXiv},
      primaryClass={cs.CL},
      url={https://arxiv.org/abs/2505.20277}, 
}

@misc{wang2024rolellmbenchmarkingelicitingenhancing,
      title={RoleLLM: Benchmarking, Eliciting, and Enhancing Role-Playing Abilities of Large Language Models}, 
      author={Zekun Moore Wang and Zhongyuan Peng and Haoran Que and Jiaheng Liu and Wangchunshu Zhou and Yuhan Wu and Hongcheng Guo and Ruitong Gan and Zehao Ni and Jian Yang and Man Zhang and Zhaoxiang Zhang and Wanli Ouyang and Ke Xu and Stephen W. Huang and Jie Fu and Junran Peng},
      year={2024},
      eprint={2310.00746},
      archivePrefix={arXiv},
      primaryClass={cs.CL},
      url={https://arxiv.org/abs/2310.00746}, 
}

@misc{li2023chatharuhirevivinganimecharacter,
      title={ChatHaruhi: Reviving Anime Character in Reality via Large Language Model}, 
      author={Cheng Li and Ziang Leng and Chenxi Yan and Junyi Shen and Hao Wang and Weishi MI and Yaying Fei and Xiaoyang Feng and Song Yan and HaoSheng Wang and Linkang Zhan and Yaokai Jia and Pingyu Wu and Haozhen Sun},
      year={2023},
      eprint={2308.09597},
      archivePrefix={arXiv},
      primaryClass={cs.CL},
      url={https://arxiv.org/abs/2308.09597}, 
}

@misc{kimiteam2025kimiaudiotechnicalreport,
      title={Kimi-Audio Technical Report}, 
      author={KimiTeam and Ding Ding and Zeqian Ju and Yichong Leng and Songxiang Liu and Tong Liu and Zeyu Shang and Kai Shen and Wei Song and Xu Tan and Heyi Tang and Zhengtao Wang and Chu Wei and Yifei Xin and Xinran Xu and Jianwei Yu and Yutao Zhang and Xinyu Zhou and Y. Charles and Jun Chen and Yanru Chen and Yulun Du and Weiran He and Zhenxing Hu and Guokun Lai and Qingcheng Li and Yangyang Liu and Weidong Sun and Jianzhou Wang and Yuzhi Wang and Yuefeng Wu and Yuxin Wu and Dongchao Yang and Hao Yang and Ying Yang and Zhilin Yang and Aoxiong Yin and Ruibin Yuan and Yutong Zhang and Zaida Zhou},
      year={2025},
      eprint={2504.18425},
      archivePrefix={arXiv},
      primaryClass={eess.AS},
      url={https://arxiv.org/abs/2504.18425}, 
}

@article{Zhang2025MiMoAudioAL,
  title={MiMo-Audio: Audio Language Models are Few-Shot Learners},
  author={Xiaomi LLM-Core Team Dong Zhang and Gang Wang and Jinlong Xue and Kai Fang and Liang Zhao and Rui Ma and Shu-Qin Ren and Shuo Liu and Tao Guo and Weiji Zhuang and Xin Zhang and Xi-Na Song and Yihan Yan and Yongzhe He and Cici and Bowen Shen and Chengxuan Zhu and Chong Ma and Chun Chen and Heyu Chen and Jiawei Li and Lei Li and Menghang Zhu and Peidian Li and Qi-ying Wang and Sirui Deng and Weimin Xiong and Wen Huang and Wenyu Yang and Yilin Jiang and Yixin Yang and Yu-Shi Tian and Yue Ma and Yue Yu and Zihan Zhang and Zihao Yue and Bangjun Xiao and Bin Xia and Bofei Gao and Bowen Ye and Can Cai and Chang Liu and Chenhong He and Chunan Li and Dawei Zhu and Duo Zhang and Fengyuan Shi and Guoan Wang and Hailin Zhang and Hanglong Lv and Hanyu Li and Hao Tian and Hengxu Qu and Hong-Mei Xu and Houbin Zhang and Huaqiu Liu and Jiangshan Duo and Jia Zuo and Jianyu Wei and Jiebao Xiao and Jinhao Dong and Jun Shi and Junhao Hu and Kainan Bao and Kang Zhou and Linghao Zhang and Meng Chen and Nuo Chen and Peng Zhang and Qian Chen and Qiantong Wang and Rang Li and Shao-yang Liu and Shengfan Wang and Shicheng Li and Shi-liang Yu and Shijie Cao and Shimao Chen and Shuhao Gu and Weikun Wang and Wen-Juan Ma and Xia Deng and Xing Yong and Xing Zhang and Xu Wang and Yi-Hao Song and Yihao Zhao and Yingbo Zhao and Yizhao Gao and Yu Cheng and Yuanfang Tu and Yudong Wang and Zhaojun Huang and Zheng-Yu Tang and Zhenrui Lin and Zhichao Song and Zhi-Yue Xu and Zhixian Zheng and Zi-Cheng Jiang},
  journal={ArXiv},
  year={2025},
  volume={abs/2512.23808},
  url={https://api.semanticscholar.org/CorpusID:284351195}
}

@article{Long2025VITAAudioFI,
  title={VITA-Audio: Fast Interleaved Cross-Modal Token Generation for Efficient Large Speech-Language Model},
  author={Zuwei Long and Yunhang Shen and Chaoyou Fu and Heting Gao and Lijiang Li and Peixian Chen and Mengdan Zhang and Hang Shao and Jian Li and Jinlong Peng and Haoyu Cao and Ke Li and Rongrong Ji and Xing Sun},
  journal={ArXiv},
  year={2025},
  volume={abs/2505.03739},
  url={https://api.semanticscholar.org/CorpusID:278339323}
}

@article{yang2024qwen2,
  title={Qwen2. 5 Technical Report},
  author={Yang, An and Yang, Baosong and Zhang, Beichen and Hui, Binyuan and Zheng, Bo and Yu, Bowen and Li, Chengyuan and Liu, Dayiheng and Huang, Fei and Wei, Haoran and others},
  journal={arXiv preprint arXiv:2412.15115},
  year={2024}
}

@article{du2024cosyvoice,
  title={Cosyvoice: A scalable multilingual zero-shot text-to-speech synthesizer based on supervised semantic tokens},
  author={Du, Zhihao and Chen, Qian and Zhang, Shiliang and Hu, Kai and Lu, Heng and Yang, Yexin and Hu, Hangrui and Zheng, Siqi and Gu, Yue and Ma, Ziyang and others},
  journal={arXiv preprint arXiv:2407.05407},
  year={2024}
}

@article{xie2024mini,
  title={Mini-omni: Language models can hear, talk while thinking in streaming},
  author={Xie, Zhifei and Wu, Changqiao},
  journal={arXiv preprint arXiv:2408.16725},
  year={2024}
}

@article{wang2024freeze,
  title={Freeze-omni: A smart and low latency speech-to-speech dialogue model with frozen llm},
  author={Wang, Xiong and Li, Yangze and Fu, Chaoyou and Xie, Lei and Li, Ke and Sun, Xing and Ma, Long},
  journal={arXiv preprint arXiv:2411.00774},
  year={2024}
}

@article{defossez2024moshi,
  title={Moshi: a speech-text foundation model for real-time dialogue},
  author={D{\'e}fossez, Alexandre and Mazar{\'e}, Laurent and Orsini, Manu and Royer, Am{\'e}lie and P{\'e}rez, Patrick and J{\'e}gou, Herv{\'e} and Grave, Edouard and Zeghidour, Neil},
  journal={arXiv preprint arXiv:2410.00037},
  year={2024}
}

@article{zeng2024glm,
  title={GLM-4-Voice: Towards Intelligent and Human-Like End-to-End Spoken Chatbot},
  author={Zeng, Aohan and Du, Zhengxiao and Liu, Mingdao and Wang, Kedong and Jiang, Shengmin and Zhao, Lei and Dong, Yuxiao and Tang, Jie},
  journal={arXiv preprint arXiv:2412.02612},
  year={2024}
}

@article{chen2025minmo,
  title={MinMo: A Multimodal Large Language Model for Seamless Voice Interaction},
  author={Chen, Qian and Chen, Yafeng and Chen, Yanni and Chen, Mengzhe and Chen, Yingda and Deng, Chong and Du, Zhihao and Gao, Ruize and Gao, Changfeng and Gao, Zhifu and others},
  journal={arXiv preprint arXiv:2501.06282},
  year={2025}
}

@article{nguyen2025spirit,
  title={Spirit-lm: Interleaved spoken and written language model},
  author={Nguyen, Tu Anh and Muller, Benjamin and Yu, Bokai and Costa-Jussa, Marta R and Elbayad, Maha and Popuri, Sravya and Ropers, Christophe and Duquenne, Paul-Ambroise and Algayres, Robin and Mavlyutov, Ruslan and others},
  journal={Transactions of the Association for Computational Linguistics},
  volume={13},
  pages={30--52},
  year={2025},
  publisher={MIT Press 255 Main Street, 9th Floor, Cambridge, Massachusetts 02142, USA~…}
}

@article{zeng2024scaling,
  title={Scaling Speech-Text Pre-training with Synthetic Interleaved Data},
  author={Zeng, Aohan and Du, Zhengxiao and Liu, Mingdao and Zhang, Lei and Jiang, Shengmin and Dong, Yuxiao and Tang, Jie},
  journal={arXiv preprint arXiv:2411.17607},
  year={2024}
}

@article{fang2024llama,
  title={Llama-omni: Seamless speech interaction with large language models},
  author={Fang, Qingkai and Guo, Shoutao and Zhou, Yan and Ma, Zhengrui and Zhang, Shaolei and Feng, Yang},
  journal={arXiv preprint arXiv:2409.06666},
  year={2024}
}

@article{chen2024slam,
  title={SLAM-Omni: Timbre-Controllable Voice Interaction System with Single-Stage Training},
  author={Chen, Wenxi and Ma, Ziyang and Yan, Ruiqi and Liang, Yuzhe and Li, Xiquan and Xu, Ruiyang and Niu, Zhikang and Zhu, Yanqiao and Yang, Yifan and Liu, Zhanxun and others},
  journal={arXiv preprint arXiv:2412.15649},
  year={2024}
}

@inproceedings{siuzdak2024snac,
  title={SNAC: Multi-Scale Neural Audio Codec},
  author={Siuzdak, Hubert and Gr{\"o}tschla, Florian and Lanzend{\"o}rfer, Luca A},
  booktitle={Audio Imagination: NeurIPS 2024 Workshop AI-Driven Speech, Music, and Sound Generation},
  year={2024}
}

@inproceedings{radford2023robust,
  title={Robust speech recognition via large-scale weak supervision},
  author={Radford, Alec and Kim, Jong Wook and Xu, Tao and Brockman, Greg and McLeavey, Christine and Sutskever, Ilya},
  booktitle={International conference on machine learning},
  pages={28492--28518},
  year={2023},
  organization={PMLR}
}

@article{gao2025lucy,
  title={Lucy: Linguistic understanding and control yielding early stage of her},
  author={Gao, Heting and Shao, Hang and Wang, Xiong and Qiu, Chaofan and Shen, Yunhang and Cai, Siqi and Shi, Yuchen and Xu, Zihan and Long, Zuwei and Zhang, Yike and others},
  journal={arXiv preprint arXiv:2501.16327},
  year={2025}
}

@article{long2025vita,
  title={VITA-Audio: Fast Interleaved Cross-Modal Token Generation for Efficient Large Speech-Language Model},
  author={Long, Zuwei and Shen, Yunhang and Fu, Chaoyou and Gao, Heting and Li, Lijiang and Chen, Peixian and Zhang, Mengdan and Shao, Hang and Li, Jian and Peng, Jinlong and others},
  journal={arXiv preprint arXiv:2505.03739},
  year={2025}
}

@article{xu2025qwen2,
  title={Qwen2. 5-omni technical report},
  author={Xu, Jin and Guo, Zhifang and He, Jinzheng and Hu, Hangrui and He, Ting and Bai, Shuai and Chen, Keqin and Wang, Jialin and Fan, Yang and Dang, Kai and others},
  journal={arXiv preprint arXiv:2503.20215},
  year={2025}
}

@misc{xu2025qwen3omnitechnicalreport,
      title={Qwen3-Omni Technical Report}, 
      author={Jin Xu and Zhifang Guo and Hangrui Hu and Yunfei Chu and Xiong Wang and Jinzheng He and Yuxuan Wang and Xian Shi and Ting He and Xinfa Zhu and Yuanjun Lv and Yongqi Wang and Dake Guo and He Wang and Linhan Ma and Pei Zhang and Xinyu Zhang and Hongkun Hao and Zishan Guo and Baosong Yang and Bin Zhang and Ziyang Ma and Xipin Wei and Shuai Bai and Keqin Chen and Xuejing Liu and Peng Wang and Mingkun Yang and Dayiheng Liu and Xingzhang Ren and Bo Zheng and Rui Men and Fan Zhou and Bowen Yu and Jianxin Yang and Le Yu and Jingren Zhou and Junyang Lin},
      year={2025},
      eprint={2509.17765},
      archivePrefix={arXiv},
      primaryClass={cs.CL},
      url={https://arxiv.org/abs/2509.17765}, 
}

@article{li2025baichuan,
  title={Baichuan-audio: A unified framework for end-to-end speech interaction},
  author={Li, Tianpeng and Liu, Jun and Zhang, Tao and Fang, Yuanbo and Pan, Da and Wang, Mingrui and Liang, Zheng and Li, Zehuan and Lin, Mingan and Dong, Guosheng and others},
  journal={arXiv preprint arXiv:2502.17239},
  year={2025}
}

@article{ding2025kimi,
  title={Kimi-audio technical report},
  author={Ding, Ding and Ju, Zeqian and Leng, Yichong and Liu, Songxiang and Liu, Tong and Shang, Zeyu and Shen, Kai and Song, Wei and Tan, Xu and Tang, Heyi and others},
  journal={arXiv preprint arXiv:2504.18425},
  year={2025}
}

@article{dai2025deep,
  title={Deep Dubbing: End-to-End Auto-Audiobook System with Text-to-Timbre and Context-Aware Instruct-TTS},
  author={Dai, Ziqi and Chen, Yiting and Xu, Jiacheng and Xie, Liufei and Wang, Yuchen and Yang, Zhenchuan and Bai, Bingsong and Gao, Yangsheng and Zhou, Wenjiang and Zhao, Weifeng and others},
  journal={arXiv preprint arXiv:2509.15845},
  year={2025}
}

@article{yang2024instructtts,
  title={Instructtts: Modelling expressive tts in discrete latent space with natural language style prompt},
  author={Yang, Dongchao and Liu, Songxiang and Huang, Rongjie and Weng, Chao and Meng, Helen},
  journal={IEEE/ACM Transactions on Audio, Speech, and Language Processing},
  volume={32},
  pages={2913--2925},
  year={2024},
  publisher={IEEE}
}

@article{du2025cosyvoice,
  title={Cosyvoice 3: Towards in-the-wild speech generation via scaling-up and post-training},
  author={Du, Zhihao and Gao, Changfeng and Wang, Yuxuan and Yu, Fan and Zhao, Tianyu and Wang, Hao and Lv, Xiang and Wang, Hui and Ni, Chongjia and Shi, Xian and others},
  journal={arXiv preprint arXiv:2505.17589},
  year={2025}
}

@article{zhou2025indextts2,
  title={IndexTTS2: A Breakthrough in Emotionally Expressive and Duration-Controlled Auto-Regressive Zero-Shot Text-to-Speech},
  author={Zhou, Siyi and Zhou, Yiquan and He, Yi and Zhou, Xun and Wang, Jinchao and Deng, Wei and Shu, Jingchen},
  journal={arXiv preprint arXiv:2506.21619},
  year={2025}
}

@article{li2025infinity,
  title={Infinity Instruct: Scaling Instruction Selection and Synthesis to Enhance Language Models},
  author={Li, Jijie and Du, Li and Zhao, Hanyu and Zhang, Bo-wen and Wang, Liangdong and Gao, Boyan and Liu, Guang and Lin, Yonghua},
  journal={arXiv preprint arXiv:2506.11116},
  year={2025}
}

@inproceedings{wang2025opens2s,
  title={Opens2s: Advancing fully open-source end-to-end empathetic large speech language model},
  author={Wang, Chen and Peng, Tianyu and Yang, Wen and Bai, Yinan and Wang, Guangfu and Lin, Jun and Jia, Lanpeng and Wu, Lingxiang and Wang, Jinqiao and Zong, Chengqing and others},
  booktitle={Proceedings of the 2025 Conference on Empirical Methods in Natural Language Processing: System Demonstrations},
  pages={906--917},
  year={2025}
}

@article{hu2026qwen3,
  title={Qwen3-TTS Technical Report},
  author={Hu, Hangrui and Zhu, Xinfa and He, Ting and Guo, Dake and Zhang, Bin and Wang, Xiong and Guo, Zhifang and Jiang, Ziyue and Hao, Hongkun and Guo, Zishan and others},
  journal={arXiv preprint arXiv:2601.15621},
  year={2026}
}

@inproceedings{lipman2023flow,
  title={Flow Matching for Generative Modeling},
  author={Lipman, Yaron and Chen, Ricky TQ and Ben-Hamu, Heli and Nickel, Maximilian and Le, Matt},
  booktitle={11th International Conference on Learning Representations, ICLR 2023},
  year={2023}
}

@article{wang2023cam++,
  title={Cam++: A fast and efficient network for speaker verification using context-aware masking},
  author={Wang, Hui and Zheng, Siqi and Chen, Yafeng and Cheng, Luyao and Chen, Qian},
  journal={arXiv preprint arXiv:2303.00332},
  year={2023}
}

@article{ma2024wenetspeech4tts,
  title={Wenetspeech4tts: A 12,800-hour mandarin tts corpus for large speech generation model benchmark},
  author={Ma, Linhan and Guo, Dake and Song, Kun and Jiang, Yuepeng and Wang, Shuai and Xue, Liumeng and Xu, Weiming and Zhao, Huan and Zhang, Binbin and Xie, Lei},
  journal={arXiv preprint arXiv:2406.05763},
  year={2024}
}

@article{zen2019libritts,
  title={Libritts: A corpus derived from librispeech for text-to-speech},
  author={Zen, Heiga and Dang, Viet and Clark, Rob and Zhang, Yu and Weiss, Ron J and Jia, Ye and Chen, Zhifeng and Wu, Yonghui},
  journal={arXiv preprint arXiv:1904.02882},
  year={2019}
}

@misc{wang2024globehighqualityenglishcorpus,
      title={GLOBE: A High-quality English Corpus with Global Accents for Zero-shot Speaker Adaptive Text-to-Speech}, 
      author={Wenbin Wang and Yang Song and Sanjay Jha},
      year={2024},
      eprint={2406.14875},
      archivePrefix={arXiv},
      primaryClass={cs.SD},
      url={https://arxiv.org/abs/2406.14875}, 
}

@inproceedings{he2024emilia,
  title={Emilia: An extensive, multilingual, and diverse speech dataset for large-scale speech generation},
  author={He, Haorui and Shang, Zengqiang and Wang, Chaoren and Li, Xuyuan and Gu, Yicheng and Hua, Hua and Liu, Liwei and Yang, Chen and Li, Jiaqi and Shi, Peiyang and others},
  booktitle={2024 IEEE Spoken Language Technology Workshop (SLT)},
  pages={885--890},
  year={2024},
  organization={IEEE}
}

@inproceedings{zhang2022wenetspeech,
  title={Wenetspeech: A 10000+ hours multi-domain mandarin corpus for speech recognition},
  author={Zhang, Binbin and Lv, Hang and Guo, Pengcheng and Shao, Qijie and Yang, Chao and Xie, Lei and Xu, Xin and Bu, Hui and Chen, Xiaoyu and Zeng, Chenchen and others},
  booktitle={ICASSP 2022-2022 IEEE International Conference on Acoustics, Speech and Signal Processing (ICASSP)},
  pages={6182--6186},
  year={2022},
  organization={IEEE}
}

@inproceedings{panayotov2015librispeech,
  title={Librispeech: an asr corpus based on public domain audio books},
  author={Panayotov, Vassil and Chen, Guoguo and Povey, Daniel and Khudanpur, Sanjeev},
  booktitle={2015 IEEE international conference on acoustics, speech and signal processing (ICASSP)},
  pages={5206--5210},
  year={2015},
  organization={IEEE}
}

@article{pratap2020mls,
  title={Mls: A large-scale multilingual dataset for speech research},
  author={Pratap, Vineel and Xu, Qiantong and Sriram, Anuroop and Synnaeve, Gabriel and Collobert, Ronan},
  journal={arXiv preprint arXiv:2012.03411},
  year={2020}
}

@inproceedings{ardila2020common,
  title={Common voice: A massively-multilingual speech corpus},
  author={Ardila, Rosana and Branson, Megan and Davis, Kelly and Kohler, Michael and Meyer, Josh and Henretty, Michael and Morais, Reuben and Saunders, Lindsay and Tyers, Francis and Weber, Gregor},
  booktitle={Proceedings of the twelfth language resources and evaluation conference},
  pages={4218--4222},
  year={2020}
}

@article{chen2021gigaspeech,
  title={Gigaspeech: An evolving, multi-domain asr corpus with 10,000 hours of transcribed audio},
  author={Chen, Guoguo and Chai, Shuzhou and Wang, Guanbo and Du, Jiayu and Zhang, Wei-Qiang and Weng, Chao and Su, Dan and Povey, Daniel and Trmal, Jan and Zhang, Junbo and others},
  journal={arXiv preprint arXiv:2106.06909},
  year={2021}
}

@article{galvez2021people,
  title={The people's speech: A large-scale diverse english speech recognition dataset for commercial usage},
  author={Galvez, Daniel and Diamos, Greg and Ciro, Juan and Cer{\'o}n, Juan Felipe and Achorn, Keith and Gopi, Anjali and Kanter, David and Lam, Maximilian and Mazumder, Mark and Reddi, Vijay Janapa},
  journal={arXiv preprint arXiv:2111.09344},
  year={2021}
}

@article{wang2021voxpopuli,
  title={VoxPopuli: A large-scale multilingual speech corpus for representation learning, semi-supervised learning and interpretation},
  author={Wang, Changhan and Riviere, Morgane and Lee, Ann and Wu, Anne and Talnikar, Chaitanya and Haziza, Daniel and Williamson, Mary and Pino, Juan and Dupoux, Emmanuel},
  journal={arXiv preprint arXiv:2101.00390},
  year={2021}
}

@inproceedings{bu2017aishell,
  title={Aishell-1: An open-source mandarin speech corpus and a speech recognition baseline},
  author={Bu, Hui and Du, Jiayu and Na, Xingyu and Wu, Bengu and Zheng, Hao},
  booktitle={2017 20th conference of the oriental chapter of the international coordinating committee on speech databases and speech I/O systems and assessment (O-COCOSDA)},
  pages={1--5},
  year={2017},
  organization={IEEE}
}

@article{du2018aishell,
  title={Aishell-2: Transforming mandarin asr research into industrial scale},
  author={Du, Jiayu and Na, Xingyu and Liu, Xuechen and Bu, Hui},
  journal={arXiv preprint arXiv:1808.10583},
  year={2018}
}

@article{shi2020aishell,
  title={Aishell-3: A multi-speaker mandarin tts corpus and the baselines},
  author={Shi, Yao and Bu, Hui and Xu, Xin and Zhang, Shaoji and Li, Ming},
  journal={arXiv preprint arXiv:2010.11567},
  year={2020}
}

@article{fu2021aishell,
  title={Aishell-4: An open source dataset for speech enhancement, separation, recognition and speaker diarization in conference scenario},
  author={Fu, Yihui and Cheng, Luyao and Lv, Shubo and Jv, Yukai and Kong, Yuxiang and Chen, Zhuo and Hu, Yanxin and Xie, Lei and Wu, Jian and Bu, Hui and others},
  journal={arXiv preprint arXiv:2104.03603},
  year={2021}
}

@inproceedings{ye2025codec,
  title={Codec does matter: Exploring the semantic shortcoming of codec for audio language model},
  author={Ye, Zhen and Sun, Peiwen and Lei, Jiahe and Lin, Hongzhan and Tan, Xu and Dai, Zheqi and Kong, Qiuqiang and Chen, Jianyi and Pan, Jiahao and Liu, Qifeng and others},
  booktitle={Proceedings of the AAAI Conference on Artificial Intelligence},
  volume={39},
  number={24},
  pages={25697--25705},
  year={2025}
}

@inproceedings{tu2024charactereval,
  title={Charactereval: A chinese benchmark for role-playing conversational agent evaluation},
  author={Tu, Quan and Fan, Shilong and Tian, Zihang and Shen, Tianhao and Shang, Shuo and Gao, Xin and Yan, Rui},
  booktitle={Proceedings of the 62nd Annual Meeting of the Association for Computational Linguistics },
  volume={1},
  number={1},
  pages={11836--11850},
  year={2024}
}

@article{yang2023baichuan,
  title={Baichuan 2: Open large-scale language models},
  author={Yang, Aiyuan and Xiao, Bin and Wang, Bingning and Zhang, Borong and Bian, Ce and Yin, Chao and Lv, Chenxu and Pan, Da and Wang, Dian and Yan, Dong and others},
  journal={arXiv preprint arXiv:2309.10305},
  year={2023}
}

@article{yang2025qwen3,
  title={Qwen3 technical report},
  author={Yang, An and Li, Anfeng and Yang, Baosong and Zhang, Beichen and Hui, Binyuan and Zheng, Bo and Yu, Bowen and Gao, Chang and Huang, Chengen and Lv, Chenxu and others},
  journal={arXiv preprint arXiv:2505.09388},
  year={2025}
}

@article{glm2024chatglm,
  title={Chatglm: A family of large language models from glm-130b to glm-4 all tools},
  author={Glm, Team and Zeng, Aohan and Xu, Bin and Wang, Bowen and Zhang, Chenhui and Yin, Da and Zhang, Dan and Rojas, Diego and Feng, Guanyu and Zhao, Hanlin and others},
  journal={arXiv preprint arXiv:2406.12793},
  year={2024}
}

@article{qwen7b,
  title={7B: Open foundation and human-aligned models (of the state-of-the-arts)},
  author={Qwen, Introducing},
  journal={URL https://github. com/QwenLM/Qwen-7B/blob/main/tech\_memo. md},
  year={2023}
}

@article{XVERSE-7B,
  title={XVERSE-7B},
  author={XVERSE-7B},
  journal={URL https://github.com/xverse-ai/XVERSE-7B/blob/main/README.md},
  year={2023}
}

@misc{qwen2025qwen25technicalreport,
      title={Qwen2.5 Technical Report}, 
      author={Qwen and : and An Yang and Baosong Yang and Beichen Zhang and Binyuan Hui and Bo Zheng and Bowen Yu and Chengyuan Li and Dayiheng Liu and Fei Huang and Haoran Wei and Huan Lin and Jian Yang and Jianhong Tu and Jianwei Zhang and Jianxin Yang and Jiaxi Yang and Jingren Zhou and Junyang Lin and Kai Dang and Keming Lu and Keqin Bao and Kexin Yang and Le Yu and Mei Li and Mingfeng Xue and Pei Zhang and Qin Zhu and Rui Men and Runji Lin and Tianhao Li and Tianyi Tang and Tingyu Xia and Xingzhang Ren and Xuancheng Ren and Yang Fan and Yang Su and Yichang Zhang and Yu Wan and Yuqiong Liu and Zeyu Cui and Zhenru Zhang and Zihan Qiu},
      year={2025},
      eprint={2412.15115},
      archivePrefix={arXiv},
      primaryClass={cs.CL},
      url={https://arxiv.org/abs/2412.15115}, 
}

@article{tang2025singmos,
  title={SingMOS-Pro: An Comprehensive Benchmark for Singing Quality Assessment},
  author={Tang, Yuxun and Liu, Lan and Feng, Wenhao and Zhao, Yiwen and Han, Jionghao and Yu, Yifeng and Shi, Jiatong and Jin, Qin},
  journal={arXiv preprint arXiv:2510.01812},
  year={2025}
}

@article{huang2024mos,
  title={Mos-bench: Benchmarking generalization abilities of subjective speech quality assessment models},
  author={Huang, Wen-Chin and Cooper, Erica and Toda, Tomoki},
  journal={arXiv preprint arXiv:2411.03715},
  year={2024}
}

@article{an2025fun,
  title={Fun-ASR Technical Report},
  author={An, Keyu and Chen, Yanni and Chen, Zhigao and Deng, Chong and Du, Zhihao and Gao, Changfeng and Gao, Zhifu and Gong, Bo and Li, Xiangang and Li, Yabin and others},
  journal={arXiv preprint arXiv:2509.12508},
  year={2025}
}
